\pdfoutput=1
\documentclass[11pt]{article}

\usepackage[preprint]{acl}

\usepackage{times}
\usepackage{latexsym}

\usepackage[T1]{fontenc}
\usepackage[utf8]{inputenc}

\usepackage{microtype}

\usepackage{inconsolata}

\usepackage{graphicx}

\usepackage{multirow}
\usepackage{booktabs}
\usepackage{makecell}
\usepackage{amsmath}
\usepackage{graphicx}
\usepackage{amsfonts}
\usepackage{amssymb}
\usepackage{fancyhdr}
\usepackage{balance}
\usepackage{xspace}
\usepackage{utfsym}
\usepackage{tabularx}
\usepackage{subcaption}
\newcommand{\paratitle}[1]{\vspace{1.5ex}\noindent\textbf{#1}}
\newcommand{\ie}{\emph{i.e.,}\xspace}

\newcommand{\eg}{\emph{e.g.,}\xspace}

\newcommand{\ignore}[1]{}

\usepackage{tcolorbox}
\definecolor{darkorange}{RGB}{255, 140, 0}
\definecolor{lightgreen}{RGB}{145, 204, 117}
\definecolor{lightyellow}{RGB}{250, 200, 88}
\definecolor{lightred}{RGB}{238, 102, 102}
\definecolor{lightblue}{RGB}{115, 192, 222}
\newtcolorbox{promptbox}[2][Prompt]{
colback=black!5!white,
arc=5pt, 
boxrule=0.5pt,
fonttitle=\bfseries,
title=#1, 
before upper={\scriptsize}, fontupper=\fontfamily{ptm}\selectfont,
colframe=#2, % 使用传递的参数来设定 colframe
}

\title{Towards Effective and Efficient Continual Pre-training of\\ Large Language Models}

\author{%
  Jie Chen, Zhipeng Chen, Jiapeng Wang, Kun Zhou, Yutao Zhu \\
  \textbf{Jinhao Jiang, Yingqian Min} \\
  \textbf{Wayne Xin Zhao, Zhicheng Dou, Jiaxin Mao, Yankai Lin, Ruihua Song, Jun Xu} \\
  \textbf{Xu Chen, Rui Yan, Zhewei Wei, Di Hu, Wenbing Huang, Ji-Rong Wen} \\
  YuLan team, Renmin University of China \\
  \texttt{\{batmanfly, dou, jrwen\}@ruc.edu.cn}
}

\begin{document}
\maketitle

\thispagestyle{fancy}
\fancyhead{}
\lhead{\raisebox{-0.08cm}{\includegraphics[height=0.6cm]{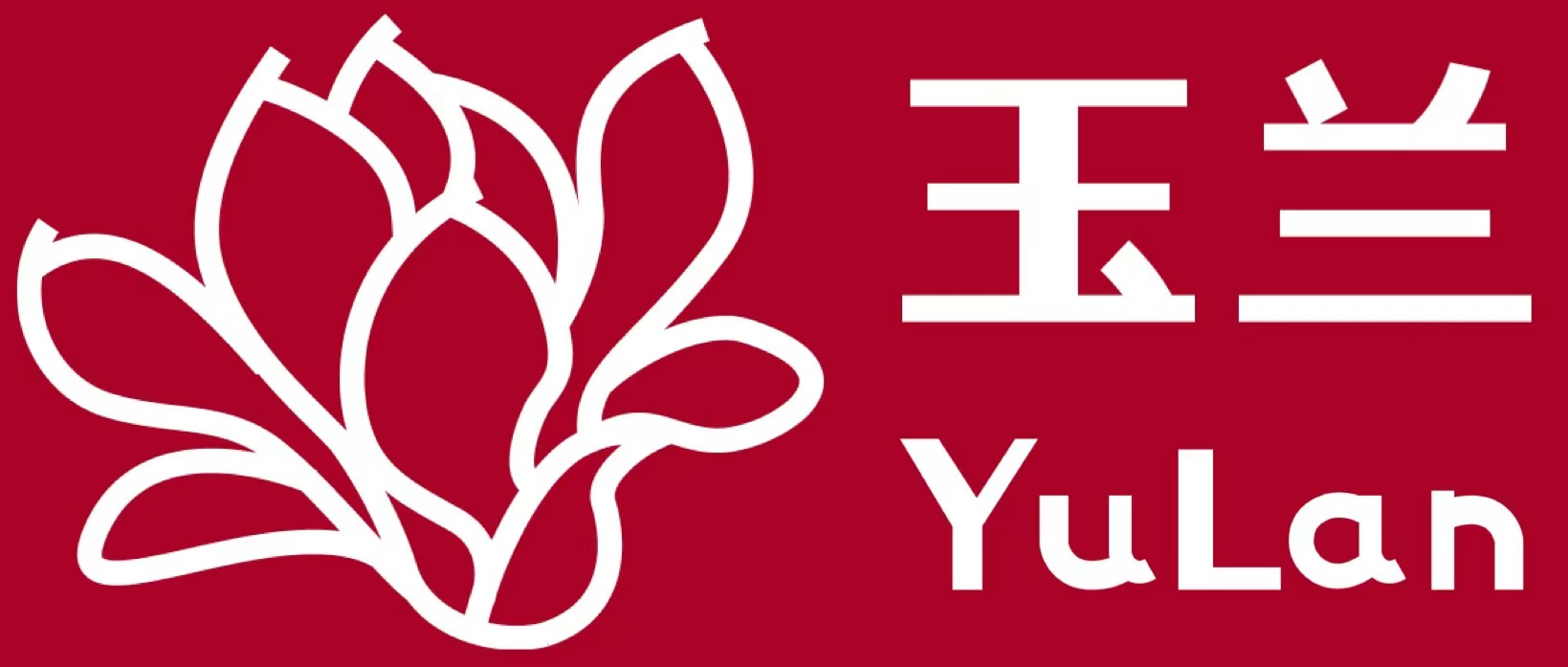}}\vspace{1mm}}

\begin{abstract}
Continual pre-training~(CPT) has been an important approach for adapting language models to specific domains or tasks. %However, most existing work either keeps training procedure unrevealed or adopts relatively simple data strategies (\eg reusing existing domain-related datasets).
To make the CPT approach more traceable, this paper presents a technical report for continually pre-training Llama-3 (8B), which significantly enhances the Chinese language ability and scientific reasoning ability of the backbone model. 
To enhance the new abilities while retaining the original abilities,  we design specific data mixture and curriculum strategies by utilizing existing datasets and synthesizing high-quality datasets. Specifically, we synthesize multidisciplinary scientific question and answer (QA) pairs based on related web pages, and subsequently incorporate these synthetic data to improve the scientific reasoning ability of Llama-3. We refer to the model after CPT as \textbf{Llama-3-SynE} (\underline{Syn}thetic data \underline{E}nhanced Llama-3).
We also present the tuning experiments with a relatively small model---TinyLlama, and employ the derived findings to train the backbone model. 
Extensive experiments on a number of evaluation benchmarks show that our approach can largely improve the performance of the backbone models, including both the general abilities (+8.81 on C-Eval and +6.31 on CMMLU) and the scientific reasoning abilities (+12.00 on MATH and +4.13 on SciEval), without hurting the original capacities.  
%Compared with increases from 16.20% to 25.40% on MATH and from 65.47% to 68.97% on SciEval.
Our model, data, and codes are available at \url{https://github.com/RUC-GSAI/Llama-3-SynE}. 
%\todo{add some results}
%Specially, we prepare detailed 
% Large language models (LLMs) have become the foundation of many applications, leveraging their extensive capabilities in processing and understanding natural language. While many open-source LLMs have been released with technical reports, the lack of training details hinders further research and development. This paper presents the development of YuLan, a series of open-source LLMs with $12$ billion parameters. The base model of YuLan is pre-trained on approximately $1.7$T tokens derived from a diverse corpus, including massive English, Chinese, and multilingual texts. We design a three-stage pre-training method to enhance YuLan's overall capabilities. Subsequent phases of training incorporate instruction-tuning and human alignment, employing a substantial volume of high-quality synthesized data. To facilitate the learning of complex and long-tail knowledge, we devise a curriculum-learning framework throughout across these stages, which helps LLMs learn knowledge in an easy-to-hard manner. YuLan's training is finished on Jan, 2024 and has achieved performance on par with state-of-the-art LLMs across various English and Chinese benchmarks. This paper outlines a comprehensive technical roadmap for developing LLMs from scratch. Our model and codes are available at \url{https://github.com/RUC-GSAI/YuLan-Chat}.
\end{abstract}

\section{Introduction}

% Llama 3 continual pre-training
% 合成数据在科学中的探索，合成题目的方法
% 合成的数据

%With the scaling of the available corpus and the success of the unsupervised pre-training method, large language models (LLMs) have achieved superior performance on various downstream scenarios and tasks~\cite{}, \eg text summarization, machine translation, and complex reasoning.
Recently, large language models (LLMs)~\cite{zhao2023survey} have achieved great progress in accelerating the development of artificial intelligence. Unlike traditional machine learning methods, LLMs basically undergo large-scale pre-training on unsupervised corpora, \eg trillions of training tokens. Through pre-training, LLMs can learn extensive knowledge from unsupervised data and acquire the capability of solving various downstream tasks via prompting~\cite{llama,gpt4,team2024gemma}. %\eg machine translation, question answering and knowledge reasoning. 
% To elicit and refine the general capacities from pre-training, 
% To further improve the capacities of LLMs, previous work collected human-annotated data to perform supervised fine-tuning (SFT)~\cite{} and reinforcement learning (RL) on LLMs~\cite{}.
% Through the further training process, LLMs can follow human instructions to generate helpful responses~\cite{}, and reduce the undesired components in the generated content (\eg hallucination or toxic content)~\cite{}.

Despite the success, LLMs still struggle in some specific scenarios, due to the large knowledge gap between pre-training data and downstream tasks. For example,  Llama-3~\cite{llama3}, primarily trained on English corpora, performs inadequately on Chinese-oriented tasks. 
Additionally, as a general-purpose LLM, Llama-3 might lack sufficient multidisciplinary scientific knowledge, \eg physics and biology.
%for science-related tasks requiring multidisciplinary knowledge (\eg mathematics, physics, and biology), \textcolor{blue}{Llama-3 is also prone to hallucinate or generate unfaithful results.} 
To address these issues, a widely-used approach is to conduct \emph{continual pre-training~(CPT)} for LLMs on specially-curated data related to the expected abilities~\cite{Ke-2023-ICLR-Continual,Gupta-2023-arxiv-Continual,Ibrahim-2024-arxiv-Simple}.
%Typically, it follows the similar way as pre-training:  first collects existing training data and then randomly samples batches for training the LLM.
% running similar training procedure as pre-training. 
During the CPT process, \emph{catastrophic forgetting}~\cite{Luo-2023-arxiv-An} has become a common technical issue, where new capabilities are improved but original capabilities are substantially hurt.  
% the primary challenge lies in enhancing new abilities while retaining existing ones. 
% during the CPT process~\cite{}. 
%To conduct effective CPT, it is key to design suitable data strategies (\ie data selection, mixture and curriculum)  so as to balance the new and old capacities. 
Although CPT has been widely conducted in existing work, the key training details (\eg data selection, mixture, and curriculum) to develop new abilities and maintain existing abilities have not been well discussed, especially how to boost the comprehensive capacities of a well-trained model under a limited training budget.  

%To solve it, a straightforward way is to include more general data into the domain-specific training corpus. Whereas, the data quality and mixture proportion are hard to adjust for balancing the forgetting and learning, and the key training details are not always fully provided.
% existing work either adopts a straightforward training approach by collecting domain-related data or keeps the key training details (\eg data mixture and curation) unrevealed. Consequently, 
%Thus, it is still unclear how to perform effective CPT under limited training budget.

In this paper, we present the technical report for continually pre-training the open-sourced LLM---Llama-3 (8B), with all experimental data, model checkpoints, and training code released.
Our focus is to enhance the model's capacities from two major aspects: \emph{Chinese language ability} and \emph{scientific reasoning ability}, while retaining its original capabilities. 
To achieve this, we design specific data curation strategies to improve the backbone models. For Chinese language ability, {we collect and select extensive Chinese text data from diverse sources for effective bilingual adaptation.} 
% \todo{to add more details (finish)}
For scientific reasoning ability, we draw inspiration from the exercises in textbooks and employ LLMs to synthesize scientific question and answer~(QA) pairs based on the content of web pages in the pre-training corpus. 
% we leverage LLMs to generate the scientific problems, based on the given pre-training corpus.
Furthermore, we also incorporate large-scale text data from various sources (\eg websites, books, and examinations) and different formats (\eg natural language and code) into the CPT data, %use the above data for continual pre-training, and also include large-scale text data from various sources (\eg website, book, and examination) and different formats (\eg natural language and code), 
to preserve the general capabilities. We carefully filter and select high-quality training data, following the processing approach used in YuLan-3~\cite{yulan}.

During the CPT process, it is key to explore various potential strategies for data collection, mixture, and curriculum design, akin to those used in standard pre-training~\cite{hu-2024-minicpm,yulan}. % By combining specific data strategies, it would lead to different pre-training datasets, and our goal is to select one suitable dataset with effective data.  
However, considering the huge experimental cost on Llama-3~(8B), we perform surrogate experiments using a relatively small model, TinyLlama~\cite{zhang-2024-tinyllama}. Based on TinyLlama, we extensively examine the effect of different data curation strategies, and further verify the findings in training Llama-3 (8B). %To maintain consistency in our model nomenclature, 
To follow the nomenclature for Llama models, we refer to the continually pre-trained model in this work as \textbf{Llama-3-SynE} (\underline{Syn}thetic data \underline{E}nhanced Llama-3).

To evaluate the effectiveness of our approach, we conduct comprehensive experiments comparing Llama-3-SynE with other competitive LLMs across various evaluation benchmarks, including general and scientific scenarios. Experimental results have shown that our data strategies significantly enhance the overall capabilities of Llama-3 (8B), particularly in Chinese language understanding and scientific knowledge reasoning.  Specifically, we find that synthetic data is very useful to enhance the capacities of LLMs in scientific knowledge reasoning. In summary, our contributions are as follows:

\begin{itemize}
    \item We present the complete training procedure for continually pre-training Llama-3 (8B), including data selection, mixture, and curriculum. Extensive experiments show that our CPT approach is very \emph{effective} ({yielding large improvements on Chinese and scientific benchmarks without hurting the performance on English benchmarks}) and \emph{efficient} ({consuming only about 100B tokens}).  
    %The proposed techniques and derived findings would be useful in various  adaptation scenarios of well trained LLMs. 
    
    \item We extensively explore the data synthesis technique, and generate high-quality scientific and code data (in the format of QA pairs). We show that these synthetic data can largely improve the corresponding capabilities of LLMs. To the best of our knowledge, it is the first public work that reports how to synthesize and utilize large-scale multidisciplinary scientific data for continual pre-training.  
    
    %test the effectiveness of synthesized multidisciplinary data during CPT. We generate scientific problem data containing 1.2 billion tokens and show that it is very useful to improve the capabilities of LLMs on scientific benchmarks.  
    \item We release the whole dataset utilized to continually pre-train Llama-3-SynE, including the general corpus comprising 98.5 billion tokens and synthetic data comprising 1.5 billion tokens focusing on scientific reasoning and coding tasks. Our dataset would be highly useful for training capable LLMs, which has been also evidenced by the surrogate model TinyLlama in our experiments. 
    
%    Extensive experiments show that our CPT approach is very \emph{effective} ({yielding large improvements on Chinese and scientific benchmarks}) and \emph{efficient} ({consuming only about 100B tokens}) in continual pre-training. 
    %Experimental results have shown that our proposed dataset is more comprehensive and has higher quality than the datasets in previous work.
\end{itemize}

%\todo{Kun: just revised the introduction. 1.Please check the name of Llama 3 and unify the name in the whole paper (finish) 2.Current version fails to report our take-away findings (especially on TinyLlama experiments); 3.Missing references (some references added); 4.Technical contribution is weak}

\section{Related Work}
In this section, we review the related work in the following three aspects.

\paragraph{Synthetic Data} 
% In order to enhance some specific abilities of LLMs~(\eg mathematical reasoning~\cite{cobbe2021gsm8k, yu2023metamath}, code generation~\cite{chen2021evaluating, Austin-arxiv-2021-Program}, and scientific QA~\cite{Sun-2024-AAAI-SciEval, Welbl-2024-EMNLP-SciQ}), it is crucial to train them with a sufficient amount of domain-related data. 
In order to enhance  specific abilities of LLMs~(\eg mathematical reasoning~\cite{cobbe2021gsm8k}, code generation~\cite{chen2021evaluating}, and scientific QA~\cite{Sun-2024-AAAI-SciEval}), it is crucial to train them with a sufficient amount of domain-related data.
However, the available real-world data may not be enough for models to acquire the necessary knowledge. To address this issue, synthetic data has been widely used in the training of LLMs, including general document data for pre-training ~\cite{Maini-2024-arxiv-Rephrasing}, instruction data for supervised fine-tuning~\cite{Xu-2023-arxiv-WizardLM}, and beyond. 
% \ignore{
There exist two primary methods for automatic data synthesis: directly prompting LLM APIs~\cite{Xu-2023-arxiv-WizardLM, Ding-2023-EMNLP-Ultrachat} and training customized synthetic models~\cite{zhou-2024-jiuzhang3, yue-2024-mammoth2}.
% There are two main automatic synthesizing methods: prompting existing LLMs and customizing synthetic models. 
By prompting with task instructions and suitable examplar data, 
 capable LLMs~(\eg GPT-4) can generate high-quality data that closely resembles the distribution of real-world data, potentially injecting the knowledge that they have acquired during training. 
% that is similar to the distribution of real-world data and may include knowledge LLMs have earned. 
% In addition, existing works also explore training customized synthetic models to save cost and generate data in more domain-specific formats or with domain-specific knowledge.
In addition, existing works also explore training relatively smaller customized  models to synthesize more domain-specific data with much less API  cost~\cite{zhou-2024-jiuzhang3}.  
% try to train customized synthetic models to generate data with more domain-specific formats or knowledge. 
% Existing works usually facilitate LLMs to automatically synthesize data having a similar distribution and format to real-world domain data. 
% \textcolor{blue}{In this work, we xxx} 
In this work, we extensively explore data synthesis technique in continual pre-training, and generate multidisciplinary scientific QA data. Our synthetic data is further utilized via   specially designed data mixture and curriculum strategies, which can effectively balance the new and original abilities of LLMs. 
%In this work, we directly explore the potential of relatively small LLM~(\eg Mistral-7B-Instruct-v0.3~\cite{jiang2023mistral}) for synthesizing the scientific instruction tuning data with further lower cost.
% }

\paragraph{Continual Pre-training} Continual pre-training, also called domain adaptive pre-training~\cite{Ke-2023-ICLR-Continual, Jang-2022-ICLR-Towards, Lesort-2021-arxiv-Understanding}, has been widely used to enhance the domain-specific abilities of a pre-trained model with new domain data. It has been a long-standing research challenge to adapt models to new domains and meanwhile prevent catastrophic forgetting~\cite{french-1999-catastrophic, Nguyen-2019-arixv-Toward}. Existing works have extensively studied fine-grained factors in mitigating catastrophic forgetting during continual pre-training, including warm-up method~\cite{Gupta-2023-arxiv-Continual}, data distribution~\cite{Ibrahim-2024-arxiv-Simple, Parmar-2024-arxiv-reuse}, and learning rate~\cite{Winata-2023-ACL-Overcoming, Scialom-2022-EMNLP-finetuned}. To address this issue, we focus on designing effective data curation strategies~(\ie data collection, mixture, and curriculum) in this work, and employ syntetic data to enhance the desired abilities of LLMs. %Notably, we utilize a small model to examine the effectiveness of our strategies on larger models, which clearly saves computational costs. 

\paragraph{Scientific Large Language Models} 
% Thanks to the excellent ability of LLMs, there emerges more and more request for LLMs to help answer scientific problems. To help LLMs understand and solve scientific problems, a lot of scientific large language models are trained by pre-training, continual pre-training, and fine-tuning.
The remarkable capabilities of LLMs have led to an increasing inclination towards their utilization in scientific application scenarios. 
% Due to the remarkable capabilities of LLMs, there is an increasing demand for their assistance in addressing scientific inquiries. 
% To facilitate the comprehension and resolution of such scientific challenges by LLMs, 
To enhance the capacity of LLMs to comprehend and resolve scientific problems,
extensive efforts have been devoted to training scientific-oriented large language models, such as mathematics LLMs~\cite{yue-2024-mammoth2, Shao-2024-arxiv-deepseekmath,zhou-2024-jiuzhang3}, biological LLMs~\cite{Timothy-2023-PoET, Zhang-2023-dnagpt} and chemical LLMs~\cite{Bagal-2022-MolGPT, Bran-2023-Chemcrow}. By leveraging the large-scale synthetic scientific QA data, our goal is to broaden the multidisciplinary scientific knowledge of LLMs while effectively preserving its original capabilities.
\section{The Proposed CPT Approach}
In this section, we present the proposed continual pre-training (CPT) approach for enhancing the \emph{Chinese} and \emph{scientific} capabilities of LLMs. 
%Our primary focus is on enhancing the Llama-3 model's general and scientific reasoning capabilities. We first describe the backbone model, Llama-3, and the surrogate model, TinyLlama, which we use for extensive exploratory experiments, and the data used in our experiments. We then detail the four strategies we have developed, including data ratio adjustment strategy, data curriculum strategy, topic-based data mixture strategy, and scientific data synthesizing strategy, followed by the introduction of our complete CPT approach.

\subsection{Overview}
We first provide an overall description of our CPT approach from three main aspects, including backbone model, data source, and training procedure. 

\paragraph{Backbone Model} To conduct the research on CPT, we adopt 
Llama-3 (8B)~\cite{llama3} as the backbone model, which has excelled in various downstream tasks such as text generation, translation, summarization, and question-answering. However, Llama-3 has been primarily pre-trained on English text data, which is inadequate in Chinese-oriented tasks. In addition, since Llama-3 was developed as a general-purpose LLM, it may also lack sufficient scientific knowledge. 
Considering these two limitations,  we aim to improve Llama-3's Chinese capacities as well as to enhance its performance in multidisciplinary scientific tasks. 
%which has been pre-trained on extensive and diverse text datasets, primarily in English.  We aim to improve Llama-3's capacities in Chinese tasks as well as to enhance its performance in multidisciplinary scientific tasks.
It is worth noting that the proposed approach can be generally applied to other backbone models, as evidenced by our experiments on the relatively smaller model TinyLlama (Section~\ref{surrogate_experiment}). 
%, we devise several CPT strategies that incorporate domain-specific data while retaining the model's original capabilities.

\paragraph{Data Source} The selection of data sources is key to the capacities of LLMs. To prepare the pre-training data, we mainly refer to the data configuration of Yulan-3~\cite{yulan}, which collects a diverse set of data, including web pages, encyclopedias, books, question-answering (QA) forums, academic papers, mathematical corpora, code, and synthetic data. Table~\ref{tab:data_stat} provides detailed information about the composition of our training data. We perform careful data cleaning following common strategies used in prior work~\cite{yulan-garden}.

% For the CPT of Llama-3 and TinyLlama, we utilize a diverse set of data, including webpages, encyclopedias, books, question-answering (QA) forums, academic papers, mathematical corpora, code and synthetic data. Table~\ref{tab:data_stat} provides detailed information about the language and volume of data. \todo{add source details} Non-synthetic data are derived from open-source pre-training datasets, while the construction of synthetic data is detailed in Section~\ref{sci_syn}.

% \begin{table}[!t]
% \centering
% \small
% \caption{Statistical information of the training corpus  for training Llama-3-SynE.}
% % \resizebox{0.49\textwidth}{!}{
% \setlength{\tabcolsep}{1.5mm}{
% \begin{tabular}{lcrr}
%     \toprule
%         \textbf{Dataset} & \textbf{Bilingual} & \textbf{Volume} & \textbf{Percentage} \\
%     \midrule
%         Web Pages & $\checkmark$ & $44.42$B & $45.33\%$ \\
%         Encyclopedia & $\checkmark$ & $4.84$B & $4.94\%$ \\
%         Books & $\checkmark$ & $15.47$B & $15.78\%$ \\
%         QA Forums & $\checkmark$ & $4.84$B & $4.94\%$ \\
%         Academic Papers & $\times$ & $7.78$B & $7.94\%$ \\
%         Mathematical Corpora & $\times$ & $7.78$B & $7.94\%$ \\
%         Code & $\times$ & $11.67$B & $11.91\%$ \\
%         Synthetic Data & $\times$ & $1.20$B & $1.22\%$ \\
%     \midrule
%         Total & - & $100.00$B  & $100.00\%$ \\
%     \bottomrule
% \end{tabular}
% }
% \label{tab:data_stat}
% \end{table}
\begin{table}[!t]
\centering
\small
\caption{Statistical information of the training corpus  for training Llama-3-SynE.}
% \resizebox{0.49\textwidth}{!}{
\setlength{\tabcolsep}{1.5mm}{
\begin{tabular}{lccr}
    \toprule
        \textbf{Dataset} & \textbf{English} & \textbf{Chinese}  &  \textbf{Volume} \\
    \midrule
        Web Pages & $\checkmark$ & $\checkmark$ & 45.18B \\
        Encyclopedia & $\checkmark$ & $\checkmark$ & 4.92B \\
        Books & $\checkmark$ & $\checkmark$ & 15.74B \\
        QA Forums & $\checkmark$ & $\checkmark$ & 4.92B \\
        Academic Papers & $\checkmark$ & $\times$ & 7.93B \\
        Mathematical Corpora & $\checkmark$ & $\times$ & 7.93B \\
        Code & $\checkmark$ & $\times$ & 11.88B \\
        Synthetic Data & $\checkmark$ & $\times$ & 1.50B \\
    \midrule
        Total & - & - & 100.00B \\
    \bottomrule
\end{tabular}
}
\label{tab:data_stat}
\end{table}

\begin{table}[!t]
    \centering
    \small
    \caption{Training data volume and strategies for the two CPT stages.}
    \label{tab:cpt_strategy}
    % \resizebox{0.49\textwidth}{!}{
    \setlength{\tabcolsep}{1mm}{
    \begin{tabular}{lccc}
        \toprule
        \multirow{2}*{\textbf{Strategy}} & \textbf{Bilingual} & \textbf{Synthetic} \\
        & \textbf{Adpatation} & \textbf{Enhancement} \\
        \midrule
        % Data Ratio Adjustment & \textit{general mixing} & \textit{synthetic mixing} \\
        Topic-based Data Mixture & $\checkmark$ & $\times$ \\
        PPL-based Data Curriculum & $\checkmark$ & $\times$ \\
        Scientific Data Synthesizing & $\times$ & $\checkmark$ \\
        Training Data Volume & 92.5B & 7.5B \\
        \bottomrule
    \end{tabular}
    }
\end{table}

\paragraph{Training Procedure} 
Overall, our training procedure consists of two main stages, namely 
\emph{bilingual adaptation stage} and \emph{synthetic enhancement stage}, which focus on improving Llama-3's Chinese and scientific capacities, respectively. 
%adapting English-oriented Llama-3 to Chinese tasks and enhancing its capability on multidisciplinary scientific tasks. 
In the CPT process, it is important to retain the original capability of Llama-3 by alleviating the effect of catastrophic forgetting. For this purpose, we design different data strategies to balance new and old abilities, which will be detailed in the following sections. The volume of training data and data strategies used in each stage are shown in Table~\ref{tab:cpt_strategy}.

% To systematically and efficiently carry out the CPT of Llama-3, we first conduct exploratory experiments using TinyLlama to optimize synthetic data proportions, data curriculum, and other factors. These preliminary experiments inform the design of our CPT strategy for Llama-3, ensuring effective and efficient training. Detailed results of these exploratory experiments are presented in Section~\ref{sec:tinyllama-experiments}. Our CPT strategy for Llama-3 is divided into two distinct phases: Stable and Synthesis. 
%\todo{add overview of two phases}

% Effective training strategies are crucial for the CPT process, enabling the model to retain its original capabilities while acquiring new domain-specific knowledge. We introduce the four 

\subsection{Bilingual Adaptation Stage} 
\label{sec:bastage}

We first introduce the training approach for improving the Chinese capacities of Llama-3. Following our previous experiences with Yulan-3~\cite{yulan}, we set the ratio of Chinese and English corpora as 2:8, to balance the Chinese and English capabilities. 
For pre-training, effective data mixture and schedule strategies are key to improving the capacities of LLMs. 
Based on the overall English-Chinese ratio, we further design two strategies to enhance knowledge learning from diverse domains or sources, namely topic-based data mixture and perplexity-based data curriculum. Next, we introduce the two techniques in detail.

\subsubsection{Topic-based Data Mixture} 
\label{topic_mix}

In prior work~\cite{Xie-nips-2023-DoReMi}, data mixture is usually conducted based on datasets or data types, \eg setting a sampling distribution to sample data instances from available datasets. In our approach, we aim to explore a more fine-grained adjustment on data mixture. % so that the semantic content of text can be considered during the mixture process.  
To achieve this goal, we consider establishing a topic taxonomy and conducting the data mixture at the topic level\footnote{In this work, ``topic'' has the similar meaning as ``category''.}.  Next, we present the topic-based data mixture method.

\begin{table}[t]
    \small
    \centering
    \begin{tabular}{ll}
        \toprule
        \textbf{Language} & \textbf{Topic} \\
        \midrule
        \multirow{10}*{English} & Mathematics and Physics \\
            & Computer Science and Engineering \\
            & Biology and Chemistry \\
            & History and Geography \\
            & Law and Policy \\
            & Philosophy and Logic \\
            & Economics and Business \\
            & Psychology and Sociology \\
            & Security and International Relations \\
            & Medicine and Health \\
               & Others \\
        \midrule
        \multirow{10}*{Chinese} & Biology and Chemistry \\
            & Computer Science and Engineering \\
            & Economics and Business \\
            & History and Geography \\
            & Law and Policy \\
            & Mathematics and Physics \\
            & Medicine and Health \\
            & Philosophy Arts and Culture \\
            & Project and Practical Management \\
            & Psychology Sociology and Education \\
               & Others \\
        \bottomrule
    \end{tabular}
    \caption{The pre-defined topics (category labels) for English and Chinese web pages, based on MMLU and CMMLU respectively. }
    \label{topic}
\end{table}

% \textcolor{blue}{topic or category?? czp: unify to topic}

%\textcolor{blue}{To enhance the model's capability in learning knowledge across diverse topics, we propose a dynamic data mixture strategy on both English and Chinese web pages, adjusting the proportions of the CPT data from the different topics. We introduce the approach to identify the topic of the web pages, track the changes of LLM performance, and adjust topic proportions as follows.(I need to revise the motivation)}

\paragraph{Topic Identification}
We train a classifier based on language models to identify the topic label for each web page. Specifically, we manually set topics as listed in Table~\ref{topic}. These topics are intentionally designed to be in alignment with the subjects of the MMLU~\cite{hendryckstest2021} and CMMLU~\cite{Li-2024-arxiv-CMMLU} benchmarks, which can also be extended to other topic taxonomies. 
Furthermore, we employ GPT-4 to annotate a small number of web pages as training data for our topic classifiers. 
Concretely, we adopt the zero-shot setting and construct the prompt by concatenating the topics and an unlabelled web page (see the prompt detail in Appendix~\ref{sec:appendix-prompt}). Then, we utilize the instructions to guide GPT-4 to annotate the unlabelled web page by these pre-determined topic labels.
%The details of the template leveraged to prompt GPT-4 are presented in Appendix~\ref{sec:appendix-prompt}. 
In order to conduct topic classification on both Chinese and English text, we train TinyBERT\footnote{\url{https://huggingface.co/huawei-noah/TinyBERT_General_4L_312D}} and BERT-Tiny-Chinese\footnote{\url{https://huggingface.co/ckiplab/bert-tiny-chinese}}  as the classifiers to identify the topic labels for English and Chinese web pages, respectively.
% (only ten topics, topics or categories?? only for Chinese?? need rewriting)
%The training data (\ie annotating the web pages with appropriate topic tags) are automatically generated by GPT-4. 
% (Specifically, we sample 2,000 data points each from Chinese and English webpages, determine the topic labels using GPT-4, and label the web data accordingly. The webpages are split into training and test dataset in a 9:1 ratio, and we train a topic classifier using TinyBERT () for English webpages and BERT-Tiny-Chinese () for Chinese webpages. The trained classifiers achieve 60\% classification accuracy on the English test dataset and 70\% classification accuracy on the Chinese test dataset.) 
With the utilization of these classifiers, the web pages can be assigned with specific topic labels.  
% During the CPT procedure, we dynamically adjust the proportions of these topic corpora based on the model's performance on the validation dataset of the corresponding topic.

\paragraph{Performance Change Tracking} 
To track the LLM's capabilities on different topic categories during the training process, we evaluate the change of the perplexity (PPL) score in each topic on the validation set. A reduction in the PPL score for a particular topic indicates an improvement in the model's capability regarding that topic. Concretely, supposing there are $n$ topics, the performance change on the $i$-th topic is:
\[
\Delta {p_i} = p_i^{(t)} - p_i^{(t-1)}, \quad i = 1, \dots, n,
\]
where $p_i^{(t)}$ and $p_i^{(t-1)}$ are the PPL on the $i$-th topic of LLM after the $t$-th and $(t-1)$-th {rounds}\footnote{A round consists of several training steps,  corresponding to the training of about 40B tokens.} of CPT process, respectively. The normalized performance change is then computed as:
\[
\delta_{p_i} = \frac{\Delta {p_i}}{\max(|\Delta {p_i}|)}, \quad i = 1, \dots, n.
\]

\paragraph{Data Mixture Adjustment}
Based on the performance change and current topic weight $w_i$ that indicates the importance of the $i$-th topic, we calculate the weight adjustment coefficient $f_i$ for training data proportions:
\[
f_i = 1 + \alpha \cdot \delta_{p_i} \cdot w_i,
\]
where $\alpha$ is a coefficient that controls the magnitude of the adjustment.
% \paragraph{Updating Data Proportions} 
After obtaining the adjustment coefficients (\ie $f_1, f_2, \dots, f_n$), we can update the data proportions for each topic based on these coefficients. 
% ensuring that the total proportion sum remains $1$. 
During training, let $r_i^{(t-1)}$ be the proportion of the $i$-th topic for the $(t-1)$-th round, then the proportion of data for the $t$-th round can be calculated as follows, 
\[
r_i^{(t)} = \frac{r_i^{(t-1)} \cdot f_i}{\sum_{j=1}^n r_j^{(t-1)} \cdot f_j}.
\]

% Through the topic-based mixture strategy introduced above, the capabilities of the LLM on different topics can be tracked and enhanced during the CPT process.
By using the topic-based mixture strategy, we can easily monitor the PPL change trend in a fine-grained way, and thus can better balance the abilities of LLMs across different topics or domains.  
%the LLM's performance degradation on specific topics can be timely discovered and enhanced in the next training round, maintaining the general abilities and balancing the different abilities of LLMs.

% \textcolor{blue}{Very confusing, topic, factor, data type, category???????}
% \todo{Add a summary sentence for the benefit of this design.}
% \todo{add appendix to reveal the detail classifier training process. add table about topic to metric of MMLU/CMMLU. optimize the strategy detail}

\subsubsection{Perplexity-based Data Curriculum} 
%For the CPT setting, we need to reserve the original  
In addition to adjusting the data mixture ratio, we also design a data curriculum strategy that organizes the training instances in a simple-to-complex manner. Curriculum learning has been demonstrated to be effective in many tasks~\cite{Bengio-2009-icml-curriculum}. Its primary principle is to gradually increase the difficulty (or complexity) of the training data. This strategy allows the model to establish a robust foundational knowledge base before learning more complex knowledge and skills. %Here, we mainly aim to reduce the effect of catastrophic forgetting when CPT incorporates training data in a different data distribution. 

Following this idea, we use the PPL score generated by the model to measure the difficulty level of the training data. 
% Our experiments in Section~\ref{surrogate_experiment} validate the effectiveness of this strategy using TinyLlama. 
% During the CPT process of Llama-3, \textcolor{blue}{we apply this strategy in the warmup and stable phase (introduced later in Section~\ref{})}. 
Training the model on Chinese text data with a progressively increasing PPL score can provide a gradual and smooth transition in training complexity. This is particularly crucial since Llama-3 is primarily trained on a large scale of English corpora with very little Chinese data. %, which would suffer from performance loss in English tasks when exposed to a substantial increase in the proportion of Chinese data.
Based on our preliminary experiments, starting with ``simpler'' Chinese data is beneficial to alleviate the performance loss (\ie catastrophic forgetting) of Llama-3 in English tasks.
% , when it is exposed to a substantial increase in the proportion of Chinese data. 
% facilitating an effective adaptation to diverse linguistic inputs.
% training the model in order of increasing PPL. This ensures a gradual and smooth transition for Llama-3 when initially encountering a significantly higher proportion of Chinese training data.
% \todo{other strategy. section ref. paratitle rename to be specific}

% This topic-based dynamic adjustment strategy ensures that the model receives balanced training across various topics, optimizing its performance on a range of evaluation metrics.

\begin{table}[t]
    \small
    \centering
    \begin{tabular}{clr}
        \toprule
        \textbf{Category} & \textbf{Discipline} & \textbf{Num. Synthetic Data} \\
        \midrule 
        \multirow{8}*{Scientific} & Mathematics & 207,448 \\ 
        & Physics & 241,516 \\ 
        & Chemistry & 30,838 \\ 
        & Biology & 25,103 \\
        & Astronomy & 24,060 \\
        & Earth Science & 7,936 \\ 
        & Medical Science & 8,199 \\ 
        & Computer Science & 475,566 \\ 
        & General Education & 572,478 \\
        \midrule
        \multirow{1}*{Code} & -  & 1,385,696 \\
        \bottomrule
    \end{tabular}
    \caption{The statistical information of the synthetic data of each discipline (in the form of QA pairs).}
    \label{discipline}
\end{table}

\subsection{Synthetic Enhancement Stage}\label{sec-synthesis}
After bilingual adaptation training, the LLM's performance on Chinese tasks can be significantly improved. In this stage, we further incorporate synthetic data to improve the multidisciplinary scientific capacities of Llama-3, inspired by prior work~\cite{zhou-2024-jiuzhang3,mixcpt-jiang-arxiv}, and the data ratio is correspondingly adjusted to 1:7:2 for Chinese, English, and synthetic data, respectively. 
Note that both the topic-based mixture strategy and perplexity-based data curriculum are no longer used in this training stage, and we randomly sample the data following the mixture proportion from the training corpus. Next, we describe our method for synthesizing data for CPT. 
% \textcolor{blue}{data mixture strategies??} 

%\subsubsection{Data Ratio}  (2) The second strategy, named the \textit{synthetic mixing}, builds upon the general mixing strategy by incorporating synthetic data. The primary goal is to enhance the model's scientific reasoning capabilities, which are predominantly in English. To accommodate the addition of synthetic data, we adjust the data language ratio to a proportion of 1:7:2 for Chinese, English, and synthetic data, respectively. 

\subsubsection{Synthesizing the Scientific QA Data}
%\label{sci_syn}
Synthetic data has been demonstrated to be effective and efficient for enhancing the capabilities of LLMs~\cite{yu2023metamath,yue2023mammoth,zhou-2024-jiuzhang3}. Following prior work~\cite{zhou-2024-jiuzhang3}, we generate synthetic data in the format of the question and answer (QA) pair, to cover a broad spectrum of multidisciplinary scientific knowledge. 
The synthetic questions and answers are concatenated into text and added to the CPT training corpora.
% synthesizing problems covering multidisciplinary science knowledge with diversity based on web pages and curated prompt.
% In our CPT approach, while the majority of our training data consists of traditional pre-training datasets, we integrate synthetic data in a question-answer (QA) format specifically designed to improve scientific reasoning capabilities without compromising existing strengths. 

Specifically, we consider nine scientific disciplines, \ie mathematics, physics, chemistry, biology, astronomy, earth science, medical science, computer science, and general education.  For each discipline, we manually collect a list of domain names relevant to the respective fields, such as \texttt{math.stackexchange.com} and \texttt{physicsforums.com}, allowing for the expansion of this list as needed to enhance the coverage.
To construct a science-related seed corpus, we collect scientific web pages from Dolma's CC~\cite{dolma} and C4~\cite{DBLP:conf/emnlp/DodgeSMAIGM021} subsets that belong to the collected domain names. 

Based on the above corpus, {we further extract the content snippets and fill in our designed prompt template}. Then, we utilize Mistral-7B-Instruct-v0.3\footnote{\url{https://huggingface.co/mistralai/Mistral-7B-Instruct-v0.3}} to generate relevant QA pairs that align with the targeted scientific discipline. These synthetic data are crafted to precisely mimic the structure and complexity of real-world scientific problems, which can enhance the model's capability for scientific problem understanding and reasoning.

\subsubsection{Synthesizing the Code QA Data}
During the preliminary experiments, we find that the coding capacities of Llama-3 are severely affected in the CPT process: sharp performance degradation is observed on the code evaluation benchmarks (\ie  HumanEval and MBPP). 
%\textcolor{blue}
%{We speculate that it is because our training data does not contain sufficient code data as that well aligns with the data format of evaluation benchmarks.} 

To retain the coding capacities of Llama-3, we adopt a similar data synthesis approach for generating high-quality code QA data. 
%For code synthetic data, sharp performance degradation for coding tasks is observed during CPT. Aiming to maintain coding abilities while improving the model's scientific capabilities, we find synthetic programming data effective mitigation.
Specifically, we expand the LeetCode dataset\footnote{\url{https://huggingface.co/datasets/greengerong/leetcode}} using the in-context learning (ICL) method. We randomly select problems from the LeetCode dataset as demonstrations, synthesize new coding problems, and generate answers for these problems. In implementation, we use Magicoder-S-DS-6.7B~\cite{Wei-2023-arxiv-magicoder} for both problems and solutions synthesis.

We present the statistical information of all synthetic data for both scientific and code in Table~\ref{discipline}. 
The details of the prompt for data synthesis and synthesis cases are provided in Appendix~\ref{sec:appendix-prompt} and~\ref{sec:appendix-case}.

\subsection{Implementation Details} 

We utilize the huggingface Transformers~\cite{wolf-2019-huggingface} to implement our experiments, using Flash Attention and DeepSpeed ZeRO Stage 2 to optimize the training efficiency. We employ AdamW optimizer~\cite{adamw} with $\beta_1=0.9$ and $\beta_2=0.95$, and use the Warmup-Stable-Decay (WSD) learning rate scheduler~\cite{hu-2024-minicpm} in the CPT process of Llama-3.
% For the CPT of Llama-3, we use the Warmup-Stable-Decay (WSD) learning rate scheduler~\cite{hu-2024-minicpm}. 
% For model warmup, we linearly increase the learning rate from $1.0 \times 10^{-7}$ to $1.0 \times 10^{-5}$. \textcolor{blue}{In the stable procedure, the learning rate remains constant at $1.0 \times 10^{-5}$. 
% Specifically, we train $10$B tokens in the warmup stage and train $100$B tokens in the stable stage.} 
For model warmup, we linearly increase the learning rate from $1.0 \times 10^{-7}$ to $1.0 \times 10^{-5}$ with $10$B tokens. {In the remaining training procedure, the learning rate remains constant at $1.0 \times 10^{-5}$.}
% \textcolor{blue}{Finally, in the Decay phase, the learning rate linearly decreases from $1.0 \times 10^{-5}$ to $5.0 \times 10^{-6}$(do you have a decay??)}. 
%For the surrogate experiments with TinyLlama, we use the same learning rate of $1.0 \times 10^{-4}$ as employed during its original pre-training. 

We conduct the CPT process using BFloat16 mixed precision, with a gradient clipping of $1.0$ to ensure training stability. To enhance computational efficiency, we apply gradient checkpointing strategy~\cite{chen-2016-training}. During training, the maximum context length is $8,192$ tokens for Llama-3.
% During training, the maximum context length is $8,192$ tokens for Llama-3 and $2,048$ tokens for TinyLlama, respectively.

% \todo{Kun: Please check if CPT is written as CTP in the whole paper.}

\section{Experiment}
In this section, we introduce the details of experiments for evaluating our approach. 

\subsection{Evaluation Benchmark}

% including the datasets utilized in the evaluation process, the baseline LLMs adopted to be compared, and the implementation details of Llama-3-SynE.

%\paratitle{Evaluation Benchmarks.} 
To ensure a comprehensive capacity assessment, we evaluate the performance of LLMs from the following aspects. 

% The statistical information about the selected evaluation benchmarks is presented in Table~\ref{}.

% \todo{add a table about the details of datasets.}

$\bullet$~\textit{Language Understanding}: We evaluate the English language understanding capability using the MMLU~\cite{hendryckstest2021}, and select CMMLU~\cite{Li-2024-arxiv-CMMLU} and C-Eval~\cite{huang2023CEval} for evaluating Chinese language understanding capability. %to test if LLMs can handle both English and Chinese languages.

$\bullet$~\textit{Coding Proficiency}: We evaluate the coding proficiency using the HumanEval~\cite{chen2021evaluating} and MBPP~\cite{Austin-arxiv-2021-Program} benchmarks, which measure the ability to generate correct code snippets based on given problems.

$\bullet$~\textit{Scientific Reasoning}: We evaluate it using several English and Chinese datasets from science and math domains, where SciQ~\cite{Welbl-2024-EMNLP-SciQ}, SciEval~\cite{Sun-2024-AAAI-SciEval}, ARC~\cite{allenai:arc} are English science reasoning datasets; SAT-Math~\cite{zhong2023agieval}, MATH~\cite{math}, GSM8K~\cite{cobbe2021gsm8k}, AQUA-RAT~\cite{ling2017program}, MAWPS~\cite{mawps}, ASDiv~\cite{Miao-2021-arixv-asdiv} are English math reasoning datasets; GaoKao~\cite{zhong2023agieval} is a Chinese benchmark including physical, chemical and mathematical reasoning subtasks.
%, which assess the models’ scientific knowledge base as well as their reasoning capabilities.

%$\bullet$~\textit{Instruction Following}: We adopt AlpacaEval 2.0~\cite{alpaca_eval} and AlignBench~\cite{liu2023alignbench} to evaluate the English and Chinese instruction following ability of LLMs, respectively.
%For AlpacaEval 2.0, we utilize \texttt{gpt-3.5-turbo} as the judge model and Llama 2 7B~\cite{llama2} as the baseline model.
%For AlignBench, we submit the LLM generated responses to the official website\footnote{\url{https://llmbench.ai/align}} for evaluation.

In order to better organize the evaluation results, we divide these evaluation benchmarks into two groups. The first group is \emph{major benchmarks}, containing MMLU, C-Eval, CMMLU, MATH, GSM8K, ASDiv, MAWPS, SAT-Math, HumanEval, and MBPP, which aim to evaluate the comprehensive capacities of LLMs. Note that we include commonly used math and code benchmarks in this group because it is standard practice to use these benchmarks for evaluating various general-purpose LLMs.  %they have been widely used for evaluating various open- or close-source LLMs. 
The second group is \emph{scientific benchmarks}, which contains SciEval, SciQ, GaoKao, ARC, and AQUA-RAT. These benchmarks have a broader coverage of multidisciplinary scientific knowledge, and they are used for evaluating the effectiveness of our data synthesis technique.  

For all the above evaluation benchmarks, we evaluate all the models using the few-shot or zero-shot settings.
Specifically, we report the eight-shot performance on GSM8K, ASDiv, and MAWPS, five-shot for C-Eval, CMMLU, MMLU, MATH, GaoKao, SciQ, SciEval, SAT-Math, and AQUA-RAT, three-shot for MBPP.
For HumanEval and ARC, we report the zero-shot evaluation performance.

% Concretely, we evaluate GSM8K, ASDiv, and MAWPS with 8-shot, C-Eval, CMMLU, MMLU, MATH, GaoKao, SciQ, SciEval with 5-shot, MBPP with 3-shot, and HumanEval and ARC with 0-shot.

\subsection{Surrogate Experiments with TinyLlama}
\label{surrogate_experiment}

% \paragraph{Surrogate Model: TinyLlama} 
Due to the significant costs involved in tuning experiments on  Llama-3~(8B), we use a relatively small model TinyLlama~\cite{zhang-2024-tinyllama} as a surrogate model for extensive exploratory experiments, and the derived findings can be employed to guide the training of Llama-3 (8B). 
% The valuable insights derived from these experiments can guide the effective CPT of Llama-3 (8B).
Specifically, TinyLlama is a language model with $1.1$ billion parameters, and it is pre-trained on three trillion tokens using the same architecture and tokenizer as Llama-2~\cite{llama2}, which is suitable for exploring the CPT strategies in our experiments. 
The implementation details of TinyLlama are similar to Llama-3, with the differences being TinyLlama’s fixed learning rate of $1.0 \times 10^{-4}$ and a maximum context length of $2,048$ tokens.
In this part, to avoid large performance discrepancies across benchmarks, for major benchmarks, we mainly select C-Eval, CMMLU, and MMLU for computing the average performance; for scientific benchmarks, we select SciEval, SciQ, and ARC for computing the average performance. We also report all benchmark results in Appendix~\ref{sec:appendix-tiny-detail}.
Next, we introduce the detailed experiments with TinyLlama. 
% designed for applications that demand minimal computational and memory resources.

% \begin{figure}[t]
%     \centering
%     \includegraphics[width=0.95\linewidth]{pic/syn_effective.pdf}
%     \caption{Performance of TinyLlama continually pre-trained on different corpora. \todo{detail description}}
%     % \vspace{-0.2cm}
%     \label{fig:syn_effective}
% \end{figure}
\begin{figure}[t]
    \centering
    \begin{subfigure}[b]{0.242\textwidth}
        \centering
        \includegraphics[width=0.95\textwidth]{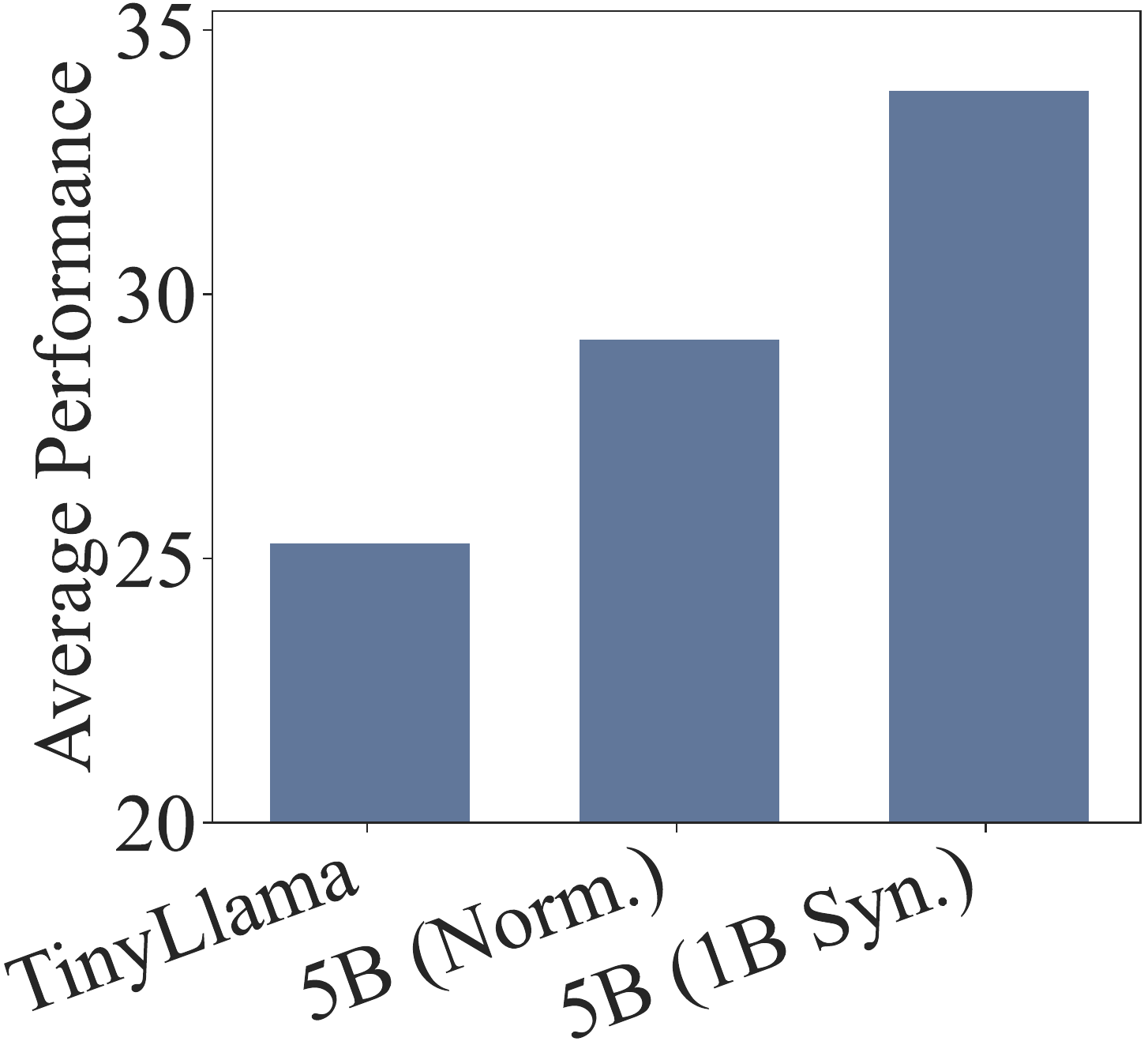}
        \caption{Major Benchmark}
        \label{fig:syn_effective_subfig1}
    \end{subfigure}
    \hspace{-0.4cm}
    \begin{subfigure}[b]{0.24\textwidth}
        \centering
        \includegraphics[width=0.95\textwidth]{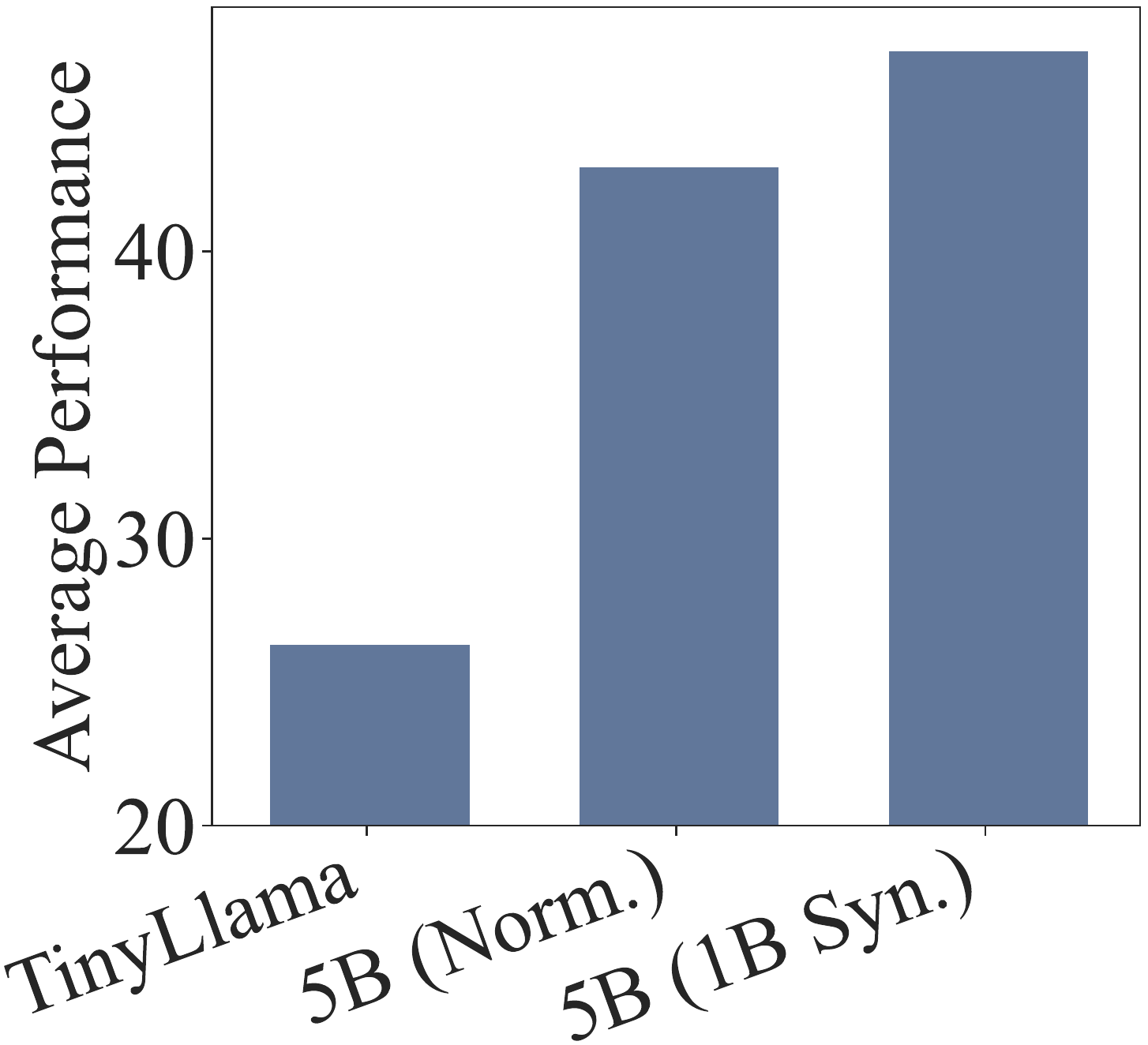}
        \caption{Scientific Benchmark}
        \label{fig:syn_effective_subfig2}
    \end{subfigure}
    \caption{Performance of TinyLlama continually pre-
trained on different corpora.}
\label{fig:syn_effective}
\end{figure}

\paragraph{Effectiveness of Synthetic Data}
To analyze the effectiveness of our CPT approach with synthetic data, we consider comparing three variants based on TinyLlama, including \emph{TinyLlama} (the original model), \emph{w/ 5B (Norm.)} (CPT with 5B normal tokens), and \emph{w/ 5B (1B Syn.)} (CPT with 4B normal tokens and 1B synthetic tokens).
In our surrogate experiments, normal training tokens are constructed by using the strategies presented in Section~\ref{sec:bastage}. 
The results are presented in Figure~\ref{fig:syn_effective}.
% we conduct ablation study on TinyLlama, and show the results in Table~\ref{tab:qualitative_analysis}.
%To study the effectiveness of the synthetic data, we implement two variations, \ie TinyLlama w/ 5b (1b sci) and w/ 5b (1b web), which continually pre-train TinyLlama on 5b data mixed with 1b synthetic science data or the original webpages.
First, by comparing with the base TinyLlama, the two variants achieve much better average performance on both major and scientific benchmarks, indicating the effectiveness our CPT data (both the collected and synthetic data). 
% we can observe that TinyLlama continually pre-trained on 5B general corpus and 1B scientific synthetic data (\ie ``w/ 5b (1b sci)'') can significantly improve the Chinese ability of LLMs, achieving much higher performance on C-Eval and CMMLU.
% Moreover, given that the scientific contents are added into the training corpus, the mathematical reasoning ability, which is also a category of scientific abilities, outperforms its backbone model (\ie TinyLlama).
% Besides, because of the obvious difference between code synthesis tasks and other tasks, the CPT process slightly hurt the code synthesis ability of LLM, which is a similar phenomenon to the Llama-3-SynE.
% Second, to assess the effectiveness of synthetic data, we adopt the scientific corpus which is utilized in generating synthetic data to replace the generated problems, and evaluate the performance of TinyLlama pre-trained on this corpus (\ie ``w/ 5b (1b web)'').
% According to the experimental results, without synthetic data in the CPT process, the performance of LLM on all downstream decreases and even performs worse than its backbone model.
Furthermore, TinyLlama \emph{w/ 5B (1B Syn.)} outperforms TinyLlama \emph{w/ 5B (Norm)}, which can demonstrate the effectiveness of our synthetic data. 
%\textcolor{blue}{It shows that our synthetic data is very useful to improve the capacities of LLMs on scientific reasoning.} 
% Since the quality of the website is difficult to control, only leveraging the original contents of these scientific websites in the continual pre-training process cannot easily guide LLMs to learn the corresponding knowledge.
Since the synthetic data is derived based on the original content of web pages, it can better extract the key knowledge of text documents and reduce the influence of irrelevant contents. Furthermore, these synthetic data are presented in the form of QA pairs, having a more similar data format with downstream tasks, which is also an important factor for performance improvement. 
%, which guarantees the data quality.

% To study the effectiveness of the synthetic data, we implement two variations, \ie TinyLlama w/ 5b (1b sci) and w/ 5b (1b web), which continually pre-train TinyLlama on 5b data mixed with 1b synthetic science data or the original webpages.
% First, by comparing with TinyLlama, the two variations both improve the performance on most benchmarks, indicating the effectiveness of continually pre-training. 

% \begin{figure}[t]
%     \centering
%     \includegraphics[width=0.98\linewidth]{pic/syn_acc.pdf}
%     \caption{Performance of TinyLlama continually pre-trained on varying levels of corrupted synthetic data. }%\todo{detail description}}
%     % \vspace{-0.2cm}
%     \label{fig:syn_acc}
% \end{figure}
\begin{figure}[t]
    \centering
    \begin{subfigure}[b]{0.241\textwidth}
        \centering
        \includegraphics[width=0.95\textwidth]{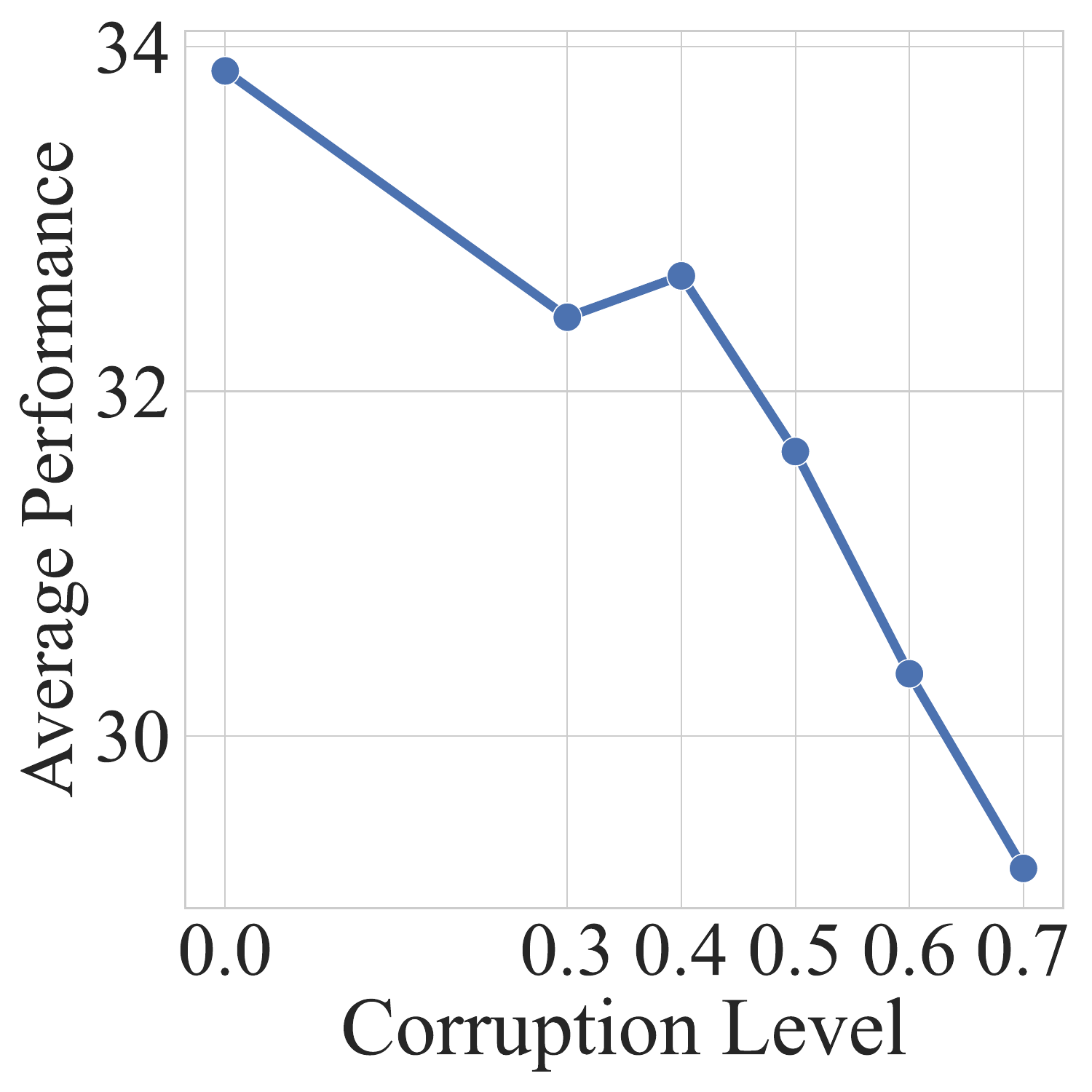}
        \caption{Major Benchmark}
        \label{fig:syn_acc_subfig1}
    \end{subfigure}
    \hspace{-0.28cm}
    \begin{subfigure}[b]{0.24\textwidth}
        \centering
        \includegraphics[width=0.95\textwidth]{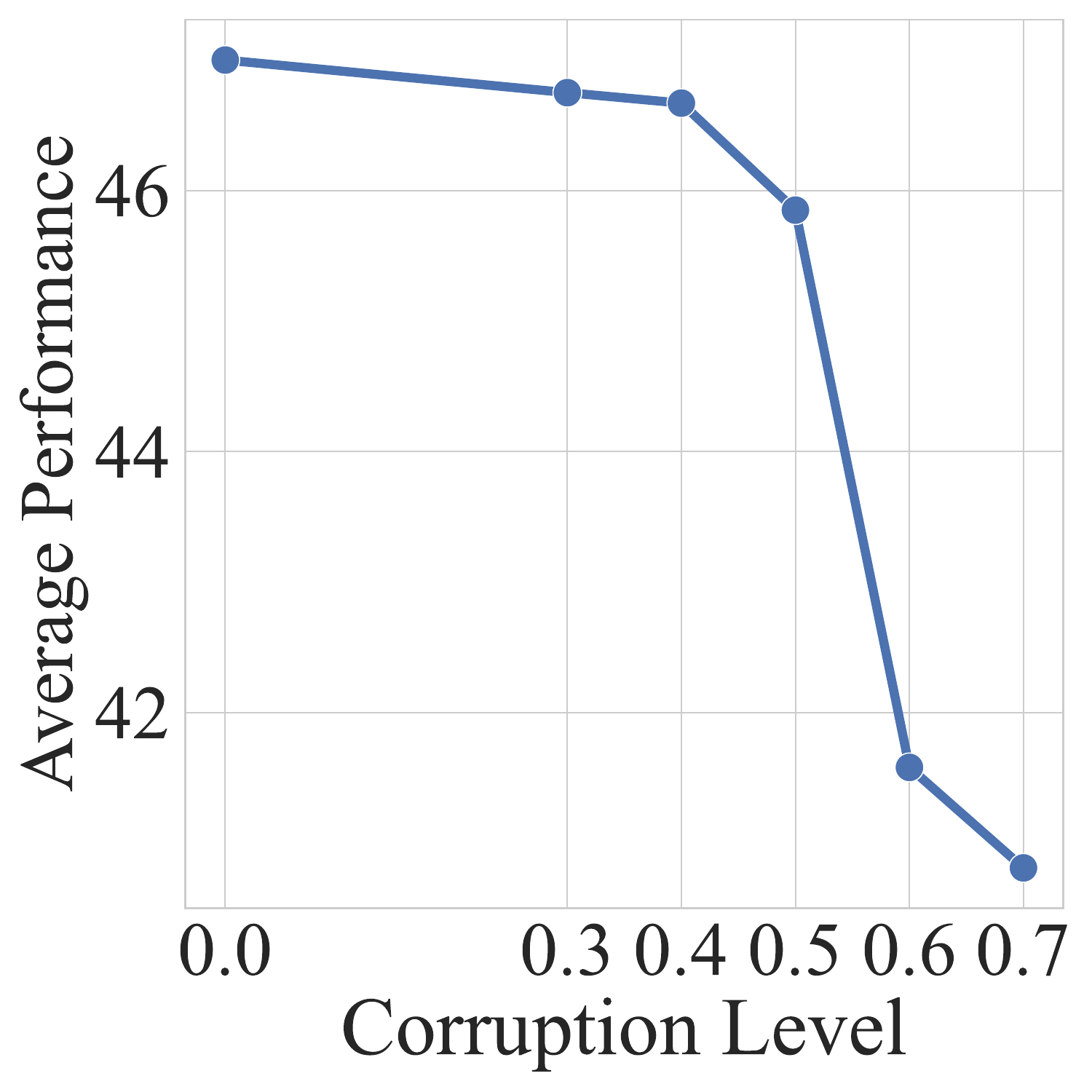}
        \caption{Scientific Benchmark}
        \label{fig:syn_acc_subfig2}
    \end{subfigure}
    \caption{Performance of TinyLlama continually pre-trained on varying corruption levels of synthetic data.}
\label{fig:syn_acc}
\end{figure}

% Does Synthetic Data Accuracy Impact Training?
\paratitle{Impact of Synthetic Data Quality} 
Intuitively, the quality (or accuracy) of synthetic data would influence the learning of domain knowledge for LLMs.
However, it is difficult to guarantee the accuracy of the automatically generated synthetic data. 
To examine the impact of the synthetic data quality, we consider simulating multiple synthetic datasets with varied 
data quality.  
% simulate the scenarios with varied data generation accuracy, we consider corrupting the generated synthetic data with different ratios. 
Concretely, we corrupt the original synthetic data  by applying 
%, creating six datasets with different degrees of accuracy degradation through the combined application of 
three types of transformation, including randomly replacing a number, substituting frequently occurring nouns with random hyponyms, and replacing frequently occurring adjectives with their antonyms (see Appendix~\ref{sec:appendix-corrupt-example}). % Examples of these transformations are provided in Appendix \ref{sec:appendix-corrupt-example}.  
%Due to the inherent limitations of synthetic models, achieving complete accuracy in synthetic data is challenging. 
% To investigate its necessity and impact, 
Based on the above transformation method, we sample one billion tokens from the synthetic data and vary the level of corruption ratios at the range of \(\{0.0, 0.3, 0.4, 0.5, 0.6, 0.7\}\).
%Concretely, the corrupt data is crafted by 
%, creating six datasets with different degrees of accuracy degradation through the combined application of 
%three types of transformations: \ie randomly replacing numbers, substituting frequently occurring nouns with random hyponyms, and replacing frequently occurring adjectives with their antonyms. Examples of these transformations are provided in Appendix \ref{sec:appendix-corrupt-example}. 
Then, we integrate 4B normal tokens with these six synthetic datasets as the CPT dataset, and train TinyLlama for performance comparison. 
% each corrupted 1B synthetic dataset is integrated with the same  corpus to form the 5B training corpus, to train TinyLlama. 
Figure~\ref{fig:syn_acc} presents the average performance of TinyLlama after training with varying corruption levels. As can be seen from this figure,  a low corruption level (\ie 0.3) has very little impact on the model performance, suggesting that LLMs can tolerate a certain degree of inaccuracy in synthetic data. However, it would still lead to  large performance degradation with a high corruption level (\ie $>0.5$). 
% high corruption levels significantly hinder model improvement. 
%Notably, even at a high corruption level of 0.7, the model's general and scientific reasoning abilities still improve after training, underscoring the robustness of our training strategy.
% These findings highlight the importance of synthetic data quality and demonstrate that while some degree of inaccuracy is tolerable, excessive corruption can adversely affect model performance.
% provide valuable insights into optimizing synthetic data for continual pre-training, ensuring a balance between data accuracy and model robustness.

% \begin{figure}[t]
%     \centering
%     \includegraphics[width=0.98\linewidth]{pic/syn_ratio.pdf}
%     \caption{Performance of TinyLlama with varying synthetic data mixing ratios.}
%     % \vspace{-0.2cm}
%     \label{fig:synth_ratio}
% \end{figure}
\begin{figure}[t]
    \centering
    \begin{subfigure}[b]{0.242\textwidth}
        \centering
        \includegraphics[width=0.95\textwidth]{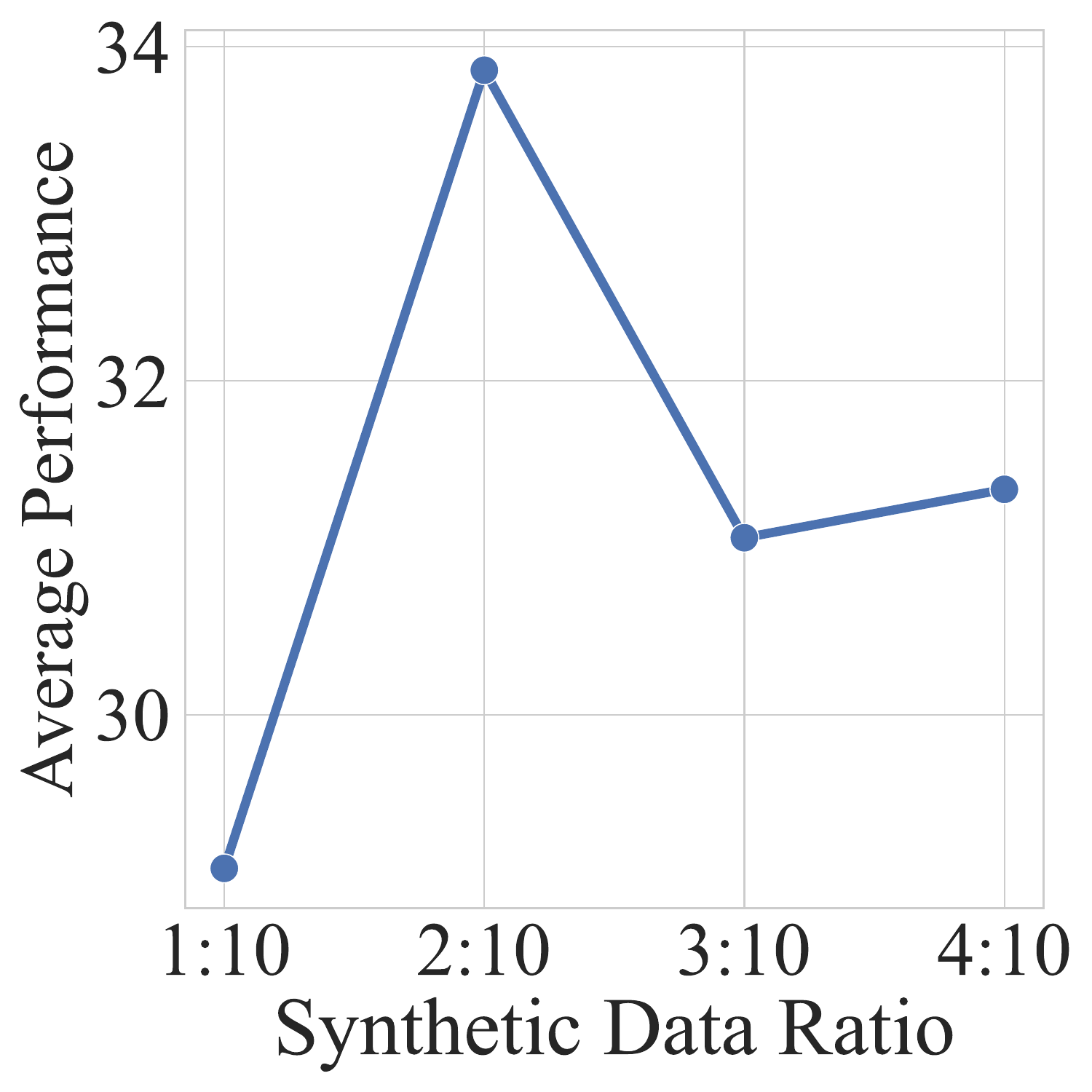}
        \caption{Major Benchmark}
        \label{fig:syn_ratio_subfig1}
    \end{subfigure}
    \hspace{-0.28cm}
    \begin{subfigure}[b]{0.24\textwidth}
        \centering
        \includegraphics[width=0.95\textwidth]{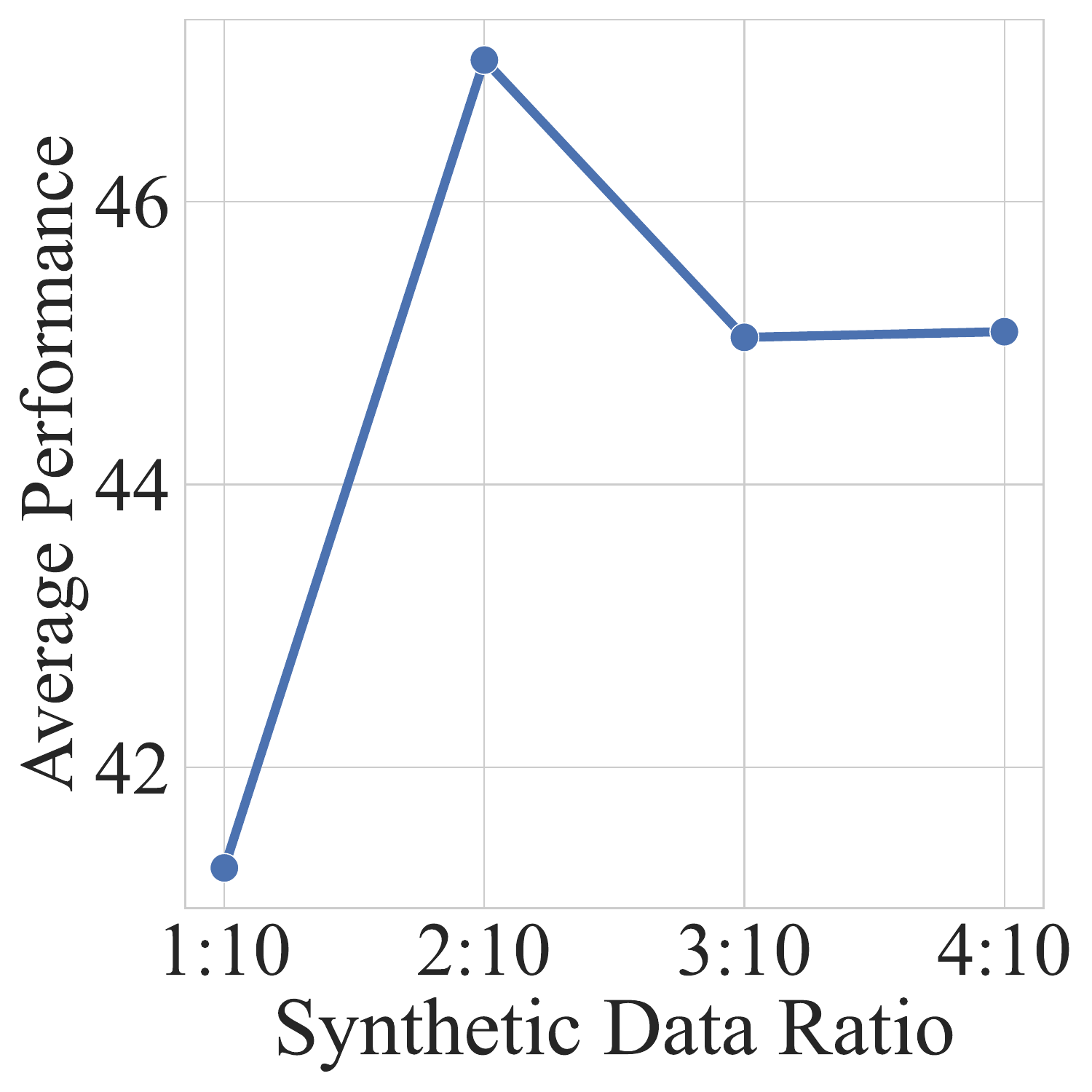}
        \caption{Scientific Benchmark}
        \label{fig:syn_ratio_subfig2}
    \end{subfigure}
    \caption{Performance of TinyLlama after training with different ratios of synthetic data.}
\label{fig:synth_ratio}
\end{figure}

\paratitle{Impact of Synthetic Data Ratio} 
For constructing the CPT dataset, we need to determine the proportion of synthetic data in the overall data distribution. 
% The data mixing ratio is crucial to balance the enhancement of new capabilities while preserving existing ones.
% , with the proportion of synthetic data potentially affecting the training outcomes. 
To investigate the effect of the mixture ratio, %we conduct experiments to explore the impact of synthetic data proportion during the Synthesis training phase. For the "Synthetic Mixing" strategy, 
we vary the  proportion of synthetic data in the training corpus, considering four choices in \(\{0.1, 0.2, 0.3, 0.4\}\), and construct a 5B-token dataset to train TinyLlama.
The relative ratios of the rest data sources are kept as that in Section~\ref{sec:bastage}. 
% Since synthetic data is mostly in English, the proportion of English data is correspondingly adjusted. 
%We construct four datasets, each containing 5 billion tokens, and use them . 
Figure~\ref{fig:synth_ratio} %\textcolor{blue}{(Table \ref{tab:science-ratio-general} and \ref{tab:science-ratio})} 
presents the average performance of TinyLlama after training with different ratios of synthetic data. 
We can see that the model's performance initially improves with the increasing of synthetic data proportions, then declines once the proportion reaches a relatively high value (\eg {40\%}). Overall, a mixture ratio of {20\%} is a good choice for integrating synthetic data and normal data. 
% \textcolor{blue}{add values}
% It indicates that more synthetic data is not necessarily better and an optimal proportion of synthetic data is beneficial, as it allows the model to effectively learn general knowledge while rapidly incorporating relevant scientific reasoning skills.

% \begin{figure}[t]
%     \centering
%     \includegraphics[width=\linewidth]{pic/data_curri.pdf}
%     \caption{Performance of TinyLlama with different data curriculum methods.}
%     % \vspace{-0.2cm}
%     \label{fig:data_curri}
% \end{figure}
\begin{figure}[t]
    \centering
    \begin{subfigure}[b]{0.24\textwidth}
        \centering
        \includegraphics[width=0.95\textwidth]{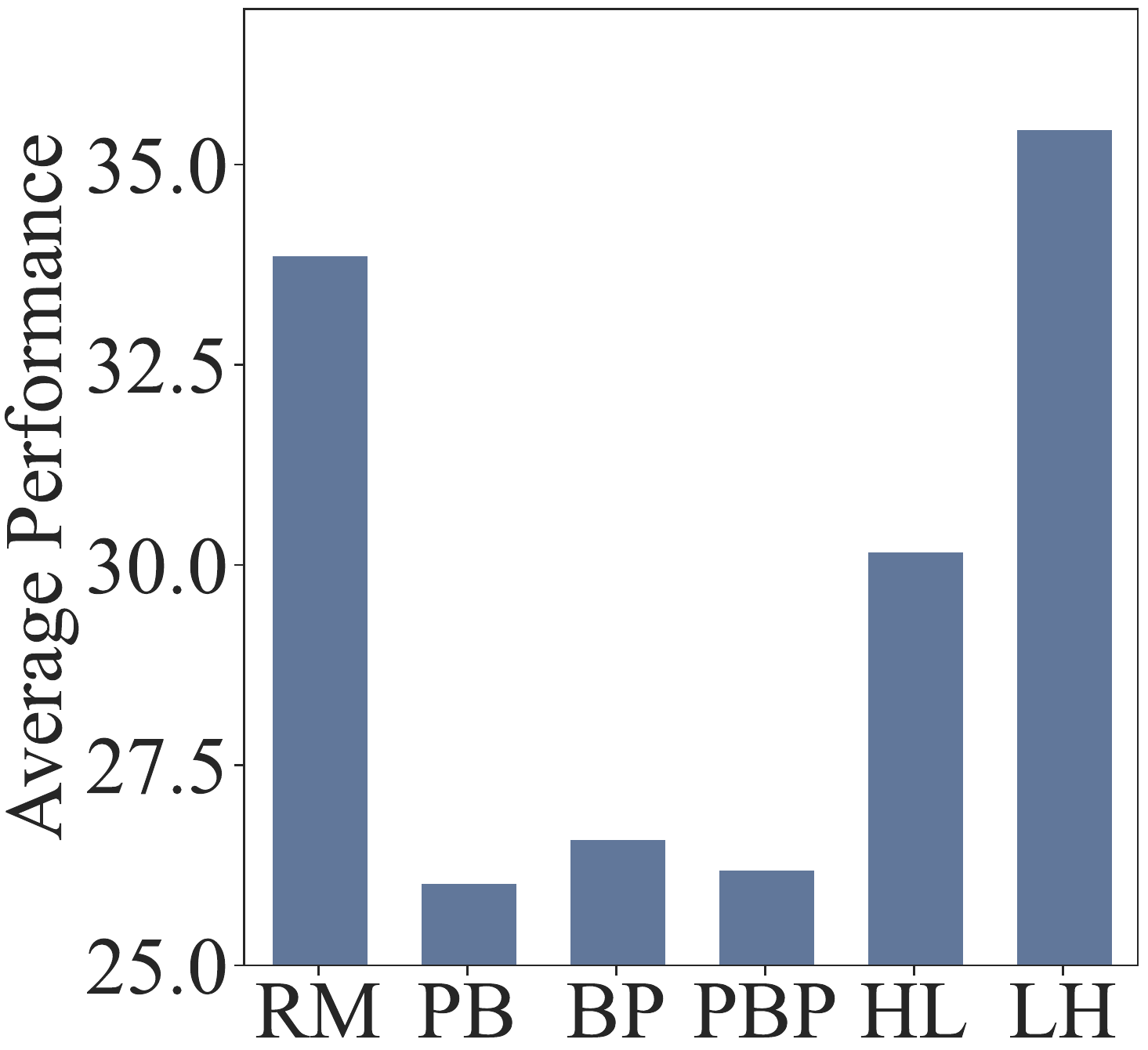}
        \caption{Major Benchmark}
        \label{fig:data_curri_subfig1}
    \end{subfigure}
    \hspace{-0.28cm}
    \begin{subfigure}[b]{0.24\textwidth}
        \centering
        \includegraphics[width=0.95\textwidth]{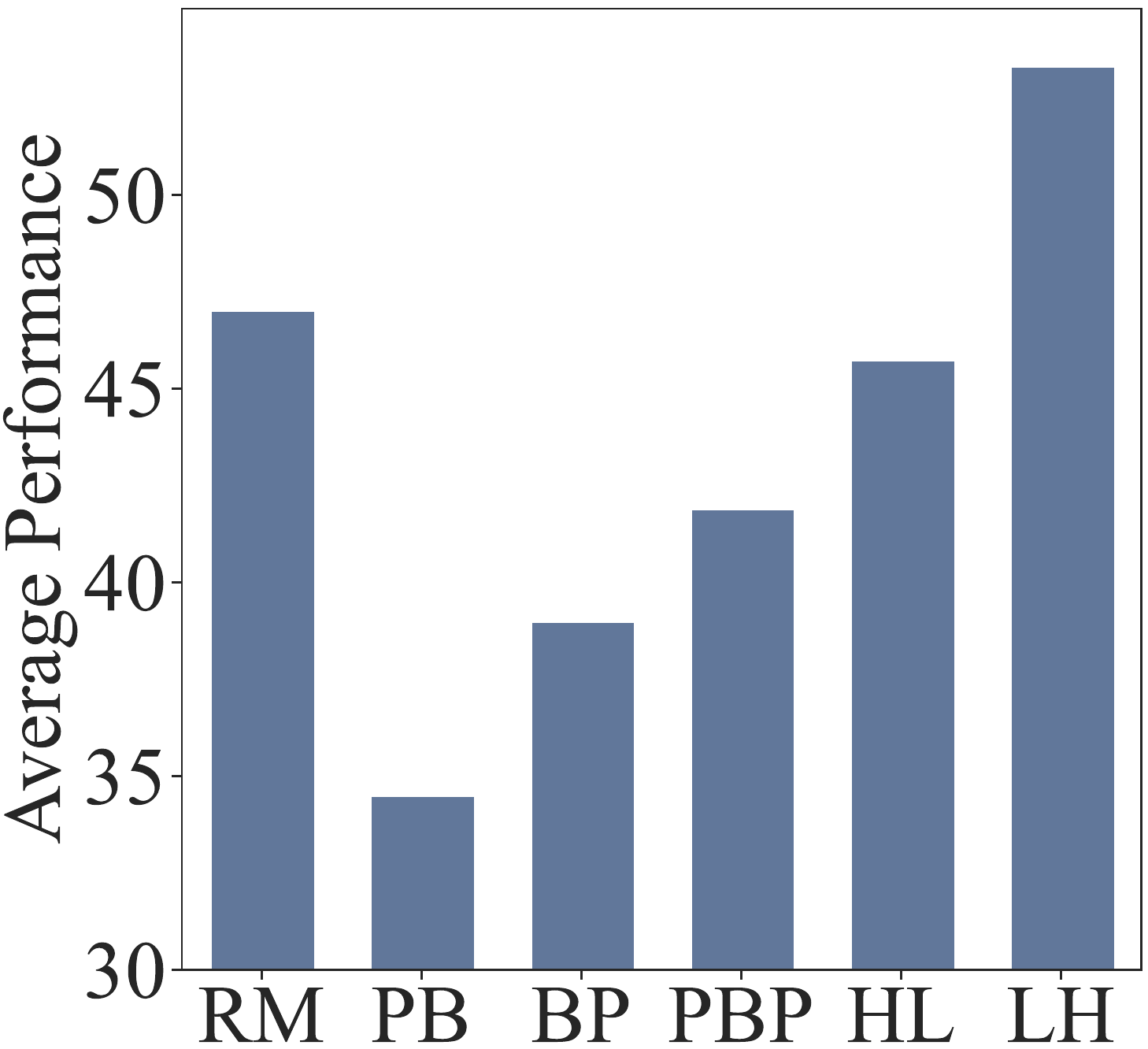}
        \caption{Scientific Benchmark}
        \label{fig:data_curri_subfig2}
    \end{subfigure}
    \caption{Performance of TinyLlama with different data curriculum methods.}
\label{fig:data_curri}
\end{figure}
% ``RM'' refers to the random mixing strategy. ``PB'', ``BP'', and ``PBP'' denote the curricula of physics $\rightarrow$ biochemistry, biochemistry $\rightarrow$ physics, and physics $\rightarrow$ biochemistry $\rightarrow$ physics, respectively. ``HL'' and ``LH'' represent the PPL reordering schedules of low $\rightarrow$ high and high $\rightarrow$ low, respectively.

\paratitle{Impact of Synthetic Data Curriculum}
In addition to the mixture ratio, we can also set different data curriculum methods (\ie reordering the instances) for synthetic data, since it mixes data from multiple disciplines. 
% A well-designed data curriculum is helpful for LLMs to learn the knowledge within the training data, which typically reorders all the instances according to a meaningful sequence, then sequentially selects the data for training.
To explore the impact of data curriculum, we consider two data instance reordering methods, either by \emph{discipline} or \emph{difficulty}, and compare these strategies with the random mixing strategy, denoted as \emph{RM}. For discipline, we design three kinds of curriculum methods by considering two disciplines, including \emph{PB} (physics $\rightarrow$ biochemistry), \emph{BP} (biochemistry $\rightarrow$ physics) and \emph{PBP} (physics $\rightarrow$  biochemistry $\rightarrow$  physics). For difficulty, we utilize the PPL score to assess the difficulty level (ten groups in total) and consider the reordering schedules of \emph{LH} (low $\rightarrow$ high) and \emph{HL} (high $\rightarrow$ low).
% To explore the impact of data curriculum, we consider two data instance reordering methods, either by \emph{discipline} or \emph{difficulty}, and compare these strategies with \emph{random mixing}. For discipline, we design three kinds of curriculum by considering two disciplines, including \emph{physics $\rightarrow$ biochemistry}, \emph{biochemistry $\rightarrow$ physics} and \emph{physics $\rightarrow$  biochemistry $\rightarrow$  physics}.
% For difficulty, we utilize the PPL score to assess the difficulty level (ten groups in total) and consider the reordering schedules of \emph{low $\rightarrow$ high} and \emph{high $\rightarrow$ low}.
%two training order: one from low to high PPL and the other from high to low PPL. 
Each data curriculum is with the same training instances but a different instance organization order.  
%the order of different domains of synthetic data and the complexity levels of the synthetic instances.  Concretely, we divide the synthetic data into three parts: physics, biochemistry, and others, and design three training courses: (1) physics-then-biochemistry; (2) biochemistry-then-physics; and (3) physics-then-biochemistry-then-physics.  To better compare the three courses, we implement them at the end of the training process. Similarly, we construct the 5B training corpus with 0.2 proportion of synthetic data, and use it to train TinyLlama. For the complexity analysis, we treat the scientific synthetic data as a whole and examine the impact of complexity levels of the synthetic data. We use PPL to assess the difficulty level and divide the scientific synthetic data into ten groups. For each group, we construct datasets containing 0.5B tokens with a 0.2 proportion of synthetic data. Finally, we design two training courses: one from low to high PPL and the other from high to low PPL. 
The results of the data curriculum are presented in Figure~\ref{fig:data_curri}.  %\textcolor{blue}{(Table \ref{tab:curriculum-general} and \ref{tab:curriculum-science})}. 
Overall, we can have  two major observations. {Firstly, the deliberate separation of data by discipline can not bring performance improvement, even hurting the model performance. %It may be because  
%The reason may be that the discipline learned before would be forgotten by the LLMs after learning the other ones.
Secondly, the easy-to-difficult curriculum can lead to more performance improvement than the contrary difficult-to-easy one and random sampling, since it can help models gradually acquire more complex knowledge information. 
This demonstrates the effectiveness of the proposed data curriculum strategy based on PPL. 
%Furthermore, the random mixing strategy performs better than the difficult-to-easy curriculum.
}

% random mixing, \textcolor{blue}{while training from difficult to easy results in the least improvement. (still improving???)} %This demonstrates the effectiveness of the data curriculum strategy based on PPL.

% may introduce distribution differences that hinder performance gains. 

% Figure () presents the average performance metrics of TinyLlama after training with the three different training sequences. The results show that while some specific data curricula are more effective than others, the improvements are less significant compared to randomly mixed training. This suggests that the deliberate separation of data by discipline may introduce distribution differences that hinder performance gains. 
% Figure () presents the average performance metrics of TinyLlama after training with these two sequences. The results indicate that the curriculum of training from easier to more difficult data improves model performance compared to random mixing, while training from difficult to easy results in the least improvement. This demonstrates the effectiveness of the data curriculum strategy based on PPL.

% \todo{disciplines order and complexity based on PPL. may divide into two parts, too long. add table, Kun: just revised the first paragraph, wanting for your added curve}

% \begin{figure}[t]
%     \centering
%     \includegraphics[width=\linewidth]{pic/syn_compare.pdf}
%     \caption{Performance of TinyLlama with different open source synthetic datasets.}
%     % \vspace{-0.2cm}
%     \label{fig:syn_compare}
% \end{figure}
\begin{figure}[t]
    \centering
    \begin{subfigure}[b]{0.242\textwidth}
        \centering
        \includegraphics[width=0.95\textwidth]{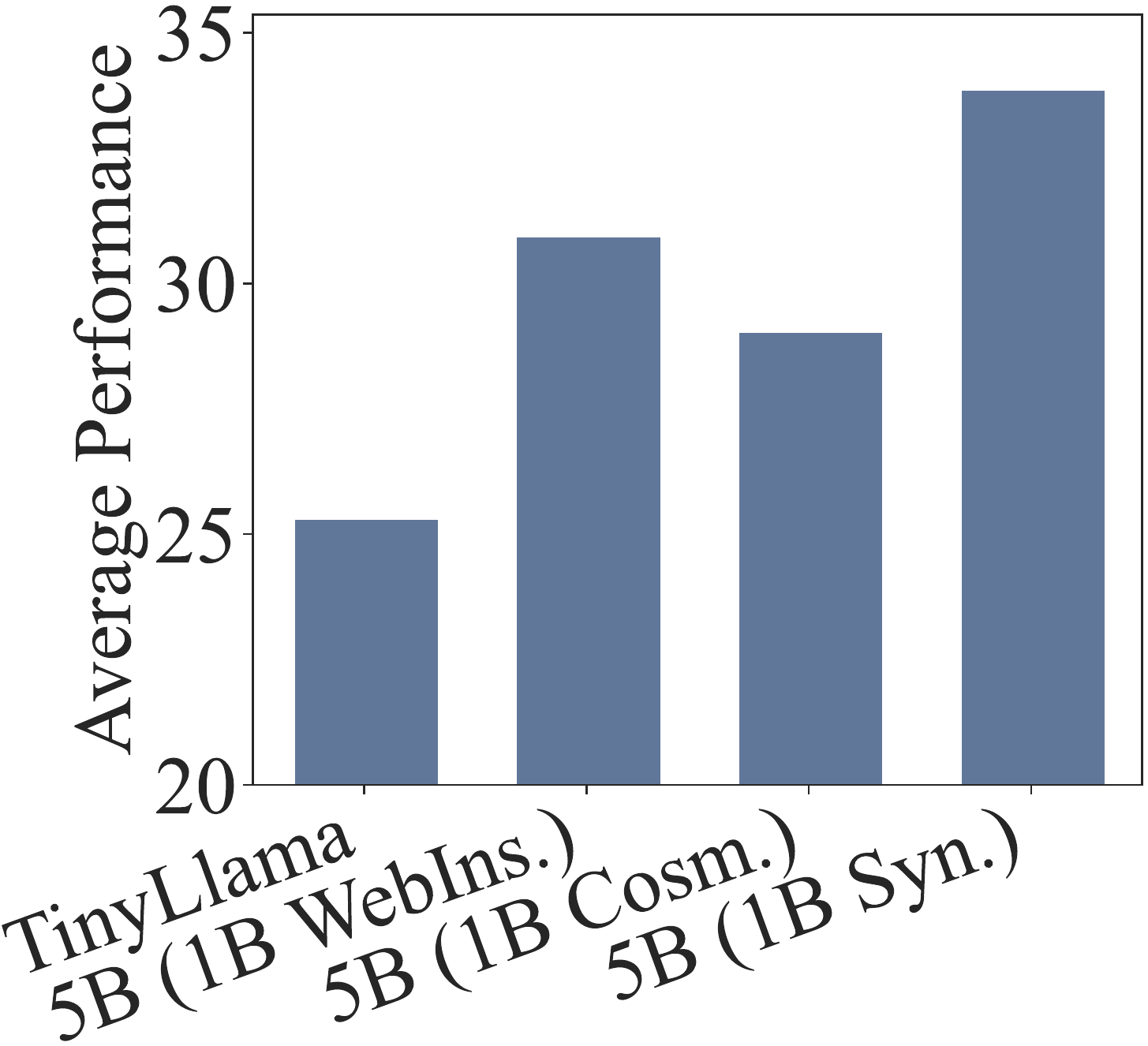}
        \caption{Major Benchmark}
        \label{fig:syn_compare_subfig1}
    \end{subfigure}
    \hspace{-0.28cm}
    \begin{subfigure}[b]{0.24\textwidth}
        \centering
        \includegraphics[width=0.95\textwidth]{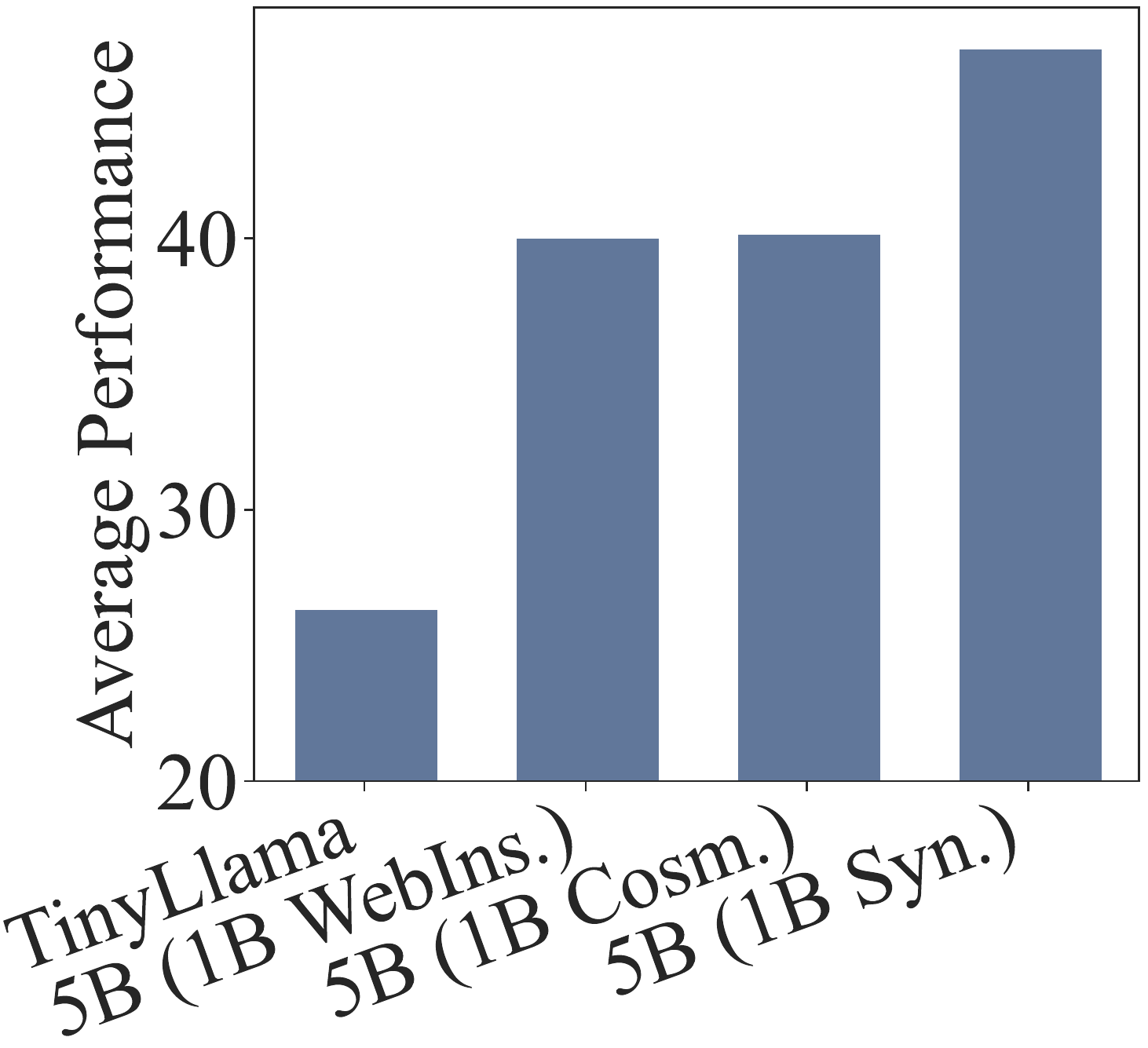}
        \caption{Scientific Benchmark}
        \label{fig:syn_compare_subfig2}
    \end{subfigure}
    \caption{Performance of TinyLlama continually pre-trained on different open-source datasets.}
\label{fig:syn_compare}
\end{figure}

\begin{table*}[t]
    \centering
    \small
    \caption{Few-shot performance comparison on major benchmarks (\ie bilingual tasks, code synthesis tasks and mathematical reasoning tasks). The best and second best are in \textbf{bold} and \underline{underlined}, respectively.}
    \setlength\tabcolsep{0.8mm}{
    \begin{tabular}{lccccccccccc}
    \toprule
    \multirow{2.5}{*}{\textbf{Models}} & \multicolumn{3}{c}{\textbf{Bilingual}} & \multicolumn{5}{c}{\textbf{Math}} & \multicolumn{2}{c}{\textbf{Code}} \\
    \cmidrule(lr){2-4}\cmidrule(lr){5-9}\cmidrule(lr){10-11}
     & \textbf{MMLU} & \textbf{C-Eval} & \textbf{CMMLU} & \textbf{MATH} & \textbf{GSM8K} & \textbf{ASDiv} & \textbf{MAWPS} & \textbf{SAT-Math} & \textbf{HumanEval} & \textbf{MBPP} \\
    \midrule[0.5pt]
    Llama-3-8B & \textbf{66.60} & 49.43 & 51.03 & 16.20 & 54.40 & 72.10 & 89.30 & 38.64 & \underline{36.59} & \textbf{47.00} \\
    DCLM-7B & 64.01 & 41.24 & 40.89 & 14.10 & 39.20 & 67.10 & 83.40 & \underline{41.36} & 21.95 & 32.60 \\
    Mistral-7B-v0.3 & 63.54 & 42.74 & 43.72 & 12.30 & 40.50 & 67.50 & 87.50 & 40.45 & 25.61 & 36.00 \\
    Llama-3-Chinese-8B & 64.10 & \underline{50.14} & \underline{51.20} & 3.60 & 0.80 & 1.90 & 0.60 & 36.82 & 9.76 & 14.80 \\
    MAmmoTH2-8B & 64.89 & 46.56 & 45.90 & \textbf{34.10} & \textbf{61.70} & \textbf{82.80} & \underline{91.50} & \underline{41.36} & 17.68 & 38.80 \\
    Galactica-6.7B & 37.13 & 26.72 & 25.53 & 5.30 & 9.60 & 40.90 & 51.70 & 23.18 & 7.31 & 2.00 \\
    \textbf{Llama-3-SynE (ours)} & \underline{65.19} & \textbf{58.24} & \textbf{57.34} & \underline{28.20} & \underline{60.80} & \underline{81.00} & \textbf{94.10} & \textbf{43.64} & \textbf{42.07} & \underline{45.60} \\
    % Mistral-7B-instruct-v0.3 & 62.34 & 43.49 & 43.56 & 13.90 & 54.00 & 73.20 & 88.70 & 39.09 & 31.70 & 37.80 \\
%     % Gemma-2-9b & 68.79 & 52.67 & 52.52 & \\
%     % SciGLM-6B & 62.11 & 63.37 & 66.28 & 23.30 & 57.60 & 72.50 & 85.90 & 24.39 & 55.40 \\
    \bottomrule
    \end{tabular}}
    \label{tab:general}
\end{table*}

\begin{table*}[t]
    \centering
    \small
    \caption{Few-shot performance comparison on scientific benchmarks. ``PHY'', ``CHE'', and ``BIO'' denote the physics, chemistry, and biology sub-tasks of the corresponding benchmarks. The best and second best are in \textbf{bold} and \underline{underlined}, respectively.}
    \setlength\tabcolsep{0.9mm}{
    \begin{tabular}{lcccccccccccc}
    \toprule
    \multirow{2.5}*{\textbf{Models}} & \multicolumn{4}{c}{\textbf{SciEval}} & \multicolumn{1}{c}{\textbf{SciQ}} & \multicolumn{3}{c}{\textbf{GaoKao}} & \multicolumn{3}{c}{\textbf{ARC}} & \multicolumn{1}{c}{\textbf{AQUA-RAT}} \\
    \cmidrule(lr){2-5}\cmidrule(lr){6-6}\cmidrule(lr){7-9}\cmidrule(lr){10-12}\cmidrule(lr){13-13}
     & {PHY} & {CHE} & {BIO} & {Avg.} & {Avg.} & {MathQA} & {CHE} & {BIO} & {Easy} & {Challenge} & {Avg.} & {Avg.} \\
    \midrule[0.5pt]
    Llama-3-8B & 46.95 & 63.45 & 74.53 & 65.47 & 90.90 & 27.92 & 32.85 & 43.81 & 91.37 & 77.73 & 84.51 & \underline{27.95} \\
    DCLM-7B & \textbf{56.71} & 64.39 & 72.03 & 66.25 & \textbf{92.50} & 29.06 & 31.40 & 37.14 & 89.52 & 76.37 & 82.94 & 20.08 \\
    Mistral-7B-v0.3 & 48.17 & 59.41 & 68.89 & 61.51 & 89.40 & 30.48 & 30.92 & 41.43 & 87.33 & 74.74 & 81.04 & 23.23 \\
    Llama-3-Chinese-8B & 48.17 & 67.34 & 73.90 & \underline{67.34} & 89.20 & 27.64 & 30.43 & 38.57 & 88.22 & 70.48 & 79.35 & 27.56 \\
    MAmmoTH2-8B & 49.39 & \textbf{69.36} & \underline{76.83} & \textbf{69.60} & 90.20 & \textbf{32.19} & \underline{36.23} & \underline{49.05} & \textbf{92.85} & \textbf{84.30} & \textbf{88.57} & 27.17 \\
    Galactica-6.7B & 34.76 & 43.39 & 54.07 & 46.27 & 71.50 & 23.65 & 27.05 & 24.76 & 65.91 & 46.76 & 56.33 & 20.87 \\
    \textbf{Llama-3-SynE (ours)} & \underline{53.66} & \underline{67.81} & \textbf{77.45} & \textbf{69.60} & \underline{91.20} & \underline{31.05} & \textbf{51.21} & \textbf{69.52} & \underline{91.58} & \underline{80.97} & \underline{86.28} & \textbf{28.74} \\
    % Gemma-2 & 25.00 & 26.91 & 32.15 & 28.62 & 26.20 & 24.22 & 26.09 & 18.57 & 22.61 & 23.64 & ~ \\
    % SciGLM-6B & ~ & ~ & ~ & ~ & ~ & 29.06 & 57.97 & 72.38 & ~ & 42.27 & ~ \\
    % Mistral-7B-instruct-v0.3 & 47.56 & 61.28 & 73.07 & 63.92 & 87.50 & 26.78 & 34.78 & 39.05 & 88.64 & 77.05 & 82.84 & 24.41 \\
    \bottomrule
    \end{tabular}
    }
    \label{tab:science}
\end{table*}

\paratitle{Comparison with Open-source Datasets} 
% The quality of pre-training data is crucial for effective CPT of LLMs. To study it, we compare our synthetic data with existing open-source synthetic pre-training datasets, to access its quality.
% Synthetic data has been widely used in existing work~\cite{yue-2024-mammoth2,zhou-2024-jiuzhang3}.
%In this work, we mainly focus on synthesizing the data that may effectively improve the specific capabilities, and would not hurt the original capability a lot.
To further examine the effectiveness of our synthetic data, we select \emph{WebInstruct} (instruction data mined from the web in the math and science domains)~\cite{yue-2024-mammoth2} and \emph{Cosmopedia} (synthetic data from the scientific subset automathtext)~\cite{benallal-2024-cosmopedia}, two large-scale open-source datasets that have been widely used for improving LLMs.
% automathtext: to improve the science knowledge of the model, we use samples from AutoMathText dataset as seed samples. The dataset covers more than just math. See this clustering plot we made.
%\textcolor{blue}{Introduce the two datasets briefly, why them}
% To further demonstrate the effectiveness, we compare the model performance using our synthetic dataset with that using existing  synthetic datasets. 
% For the fair comparison, we construct a 5B-token corpus with different datasets in the same data proportion.
For the fair comparison, we consider comparing four variants based on TinyLlama, including \emph{TinyLlama} (the original model), \emph{w/ 5B (1B Webins.)} (CPT with 4B normal tokens and 1B WebInstruct tokens), \emph{w/ 5B (1B Cosm.)} (CPT with 4B normal tokens and 1B Cosmopedia tokens), and \emph{w/ 5B (1B Syn.)} (CPT with 4B normal tokens and 1B tokens from our synthetic data).
% Similarly, we construct two 5B corpus by replacing the synthetic data with the open-source one and train TinyLlama. 
Figure~\ref{fig:syn_compare}  %\textcolor{blue}{(Table \ref{tab:compare-dataset-general} and \ref{tab:compare-dataset-science})} 
presents the performance of TinyLlama after training with different open-source datasets. The results show that our synthetic data leads to more improvements in both major and scientific benchmarks, which demonstrates the effectiveness of our data synthesis method. %It indicates that our synthetic datasets are also high-quality ones and can be more suitable for scientific reasoning capabilities.

% \begin{table}[t]
%     \centering
%     \small
%     \caption{Performance of SFT LLMs on the instruction-following benchmarks. The best and second best are in \textbf{bold} and \underline{underlined}, respectively.}
%     \todo{omit sft results}
%     \setlength\tabcolsep{1mm}{
%     \begin{tabular}{lcc}
%     \toprule
%     \textbf{Models} & \textbf{AlpacaEval 2.0} & \textbf{AlignBench} \\
%     \midrule
%     Llama-3-8b &  &  \\
%     Llama-3-SynE-SFT & 43.03 & 4.67 \\
%     Llama-3-SynE-RLHF & 43.85 & 5.27 \\
%     \bottomrule
%     \end{tabular}}
%     \label{tab:sft}
% \end{table}

%\paragraph{Effectiveness of Data Mixture Strategy.}
%balance the abilities of the general domain and the specific domain and reduce the overfitting of LLMs to the specific downstream scenarios.
%With the removal of the data mixture strategy, we can observe that synthetic data (\ie ``w/ 5b rand (1b sci)'') also expresses its effectiveness and outperforms the model trained on the original scientific websites (\ie ``w/ 5b rand (1b web)'').
%These results have verified the effectiveness of the data mixture strategy and the synthetic data.

%\todo{Kun: just revised, should it remove into the 4.2 section as all the experiments are based on TinyLlama.}

\subsection{Main Experiments with Llama-3}
% What Did We Learn from TinyLlama for Llama-3?
% How Did We Continue Pre-training Llama-3?

Based on the above findings from TinyLlama, we adopt the best-performing strategies or configurations for continual pre-training Llama-3. %, to enhance its capacities.
%of Llama-3-SynE, and conduct the experiments to assess the performance of Llama-3-SynE and the transferability of the conclusions from TinyLlama.

\paratitle{Baselines} 
To conduct the comprehensive evaluation, we adopt both general LLMs and scientific LLMs as baselines in our experiment.
We consider three kinds of LLMs as baselines, including \emph{general-purpose LLM}, \emph{scientific LLM} (enhanced by the science-related corpus or instructions), and \emph{continual pre-training LLM}.  
%General LLMs denote the LLMs pre-trained and supervised fine-tuned on the general corpus and instructions, without improving the abilities of specific downstream scenarios.
For general-purpose LLMs, we adopt DCLM-7B~\cite{Li-2024-arxiv-datacomp} and Mistral-7B-v0.3~\cite{jiang2023mistral} as the baseline in the evaluation. 
%cientific LLMs refer to the LLMs trained on the science-related corpus or instructions to enhance their scientific capacities.
For scientific LLMs, we adopt MAmmoTH2-8B~\cite{yue-2024-mammoth2} and Galactica-6.7B~\cite{taylor-2022-galactica} as the baseline LLMs.
In addition, we also report the evaluation results of Llama-3-Chinese-8B\footnote{\url{https://huggingface.co/hfl/llama-3-chinese-8b}}, which has also been continually pre-trained based on Llama-3.

% the Llama3-SynE intermediate checkpoints during the pre-training process. 
% We adopt 
% \textcolor{blue}{contintual pre-trianing baselines....}

%\todo{Kun: deciding the baselines ASAP please, then I will revise it.}

\paragraph{Results on Major Benchmarks} 
As presented in Table~\ref{tab:general}, we can observe that Llama-3-SynE outperforms its backbone model Llama-3 (8B) by a large margin on Chinese evaluation benchmarks (\eg C-Eval and CMMLU). It shows that our approach is very effective for enhancing the Chinese language capacity of Llama-3. We carefully collect and clean the Chinese text data, and also design suitable data mixture and curriculum to adaptively retrain these models, which is the key to performance improvement on Chinese benchmarks. 
%, and it also achieves improved or comparable performance on the rest benchmarks (\eg MMLU). 
%It indicates the effectiveness of our CPT approach for enhancing the capacity of Chinese language understanding.
%Given that the Chinese corpus is in a low proportion to the Llama-3 pre-training contents, improving the ratio of Chinese corpus in the pre-training dataset can significantly enhance the Chinese abilities of Llama-3-SynE.
%It is interesting to find that only a small amount of Chinese corpus (xxxB) can largely improve the Chinese ability, which indicates that different languages share similar knowledge, and the high-quality corpus of a specific language can activate the ability of LLMs in the corresponding language.
Second, for English evaluation benchmarks, our approach slightly underperforms Llama-3~(8B) on MMLU, while achieving improved or comparable performance on the rest math and code benchmarks. 
It demonstrates that our approach can well address the catastrophic forgetting issue of the original capabilities of LLMs.
Actually, based on our preliminary experiments (also evidenced by baseline models), Chinese-adaptive CPT models are difficult to retain the original performance on English-oriented benchmarks (\eg MMLU) due to the data distribution discrepancy between pre-training and CPT. %the potential inconsistencies between pre-training data and CPT data.  
These results indicate that our approach can effectively balance the original and new capacities. 
%The reason behind this is the adoption of our .
%During the pre-training process, the general abilities might be hurt because of the improved ratio of Chinese training content.
%Once the training stage ends, the proportion of different domains is changed, maintaining and recovering the corresponding capacities of Llama-3-SynE.
%Third, the ability of Llama-3-SynE to synthesize code on the given problem decreases, because we focus on enhancing the bilingual and scientific ability through CPT, and do not consider the code synthesis capacity, which has a larger gap with the skills concerned in this work.
%\textcolor{blue}{Furthermore, by comparing Llama-3-SynE with other science-specific LLMs (\eg Galactica-6.7B and SciGLM-6B), we can observe that Llama-3-SynE can achieve comparable performance with them and even performs better in few tasks. The reason is that we can dynamically adjust the proportion of  training corpus, which enables us to well balance the general and specific skills during CPT.} 

%\todo{Kun: just revised, requiring to fill the blanks and decide if we should delete the code benchmark with negative results.}

\paragraph{Results on Scientific Benchmarks} 
As shown in Table~\ref{tab:science}, Llama-3-SynE performs very well on the scientific benchmarks, which is consistently better than the backbone model Llama-3.
It indicates that our synthetic data is very effective in improving the scientific reasoning capability of LLMs.
%the training efficiency has been improved, \eg the higher utilization of unlabeled corpus and better performance on evaluation benchmarks.
In particular, compared to the English datasets, Llama-3-SynE achieves a significantly larger improvement on the Chinese datasets, \ie GaoKao BIO benchmark (25.71 points improvement over Llama-3), since our CPT model can effectively balance the English and Chinese reasoning abilities on scientific tasks.  
Among all the baselines, MAmmoTH2-8B achieves very good performance on English scientific benchmarks, while it suffers from performance degradation on general Chinese benchmarks, \eg C-Eval and CMMLU. 

By combining the results on major and scientific benchmarks, we can see that Llama-3-SynE  achieves very competitive performance in various abilities, and it can effectively alleviate the catastrophic forgetting issue in the CPT process. Our CPT approach only consumes %\textcolor{blue}
{about 100B tokens}, which is relatively efficient in training compute.

\section{Conclusion}
In this work, we studied how to perform effective continual pre-training~(CPT) for LLMs under a limited training budget. Our focus is to develop new capabilities and meanwhile avoid catastrophic forgetting of original capabilities. Specifically, we extensively explored the data synthesis technique, and generated high-quality scientific and code data, which can largely improve the corresponding abilities of LLMs. 
In order to reduce the tuning cost, 
we conducted extensive experiments on TinyLlama by examining various data curation strategies, including data selection, mixture, and curriculum.
The derived findings were further employed to guide the training of Llama-3 (8B). 
%Then, we test the performance of all the trained TinyLlamas on Chinese language understanding and multidisciplinary science knowledge benchmarks, to obtain the best-performed recipe for training Llama-3 (8B), namely Llama3-zh-sci.
Experimental results have shown that our CPT approach can largely boost the Chinese and scientific reasoning abilities of the backbone model, and meanwhile effectively retain its original abilities. 
% can perform better than baselines on the benchmarks about expected and original capabilities.

%In the future, we will conduct more experiments to study the effect of other influencing factors in the CPT process. 
%Besides, we will investigate if the CPT recipe can be transferred into other advanced capabilities and low-resource languages.
% \input{sections/limitations}

\section*{Acknowledgment}
The computing resources are supported by Public Computing Cloud, Renmin University of China.

% \clearpage

\bibliography{main}

\balance
\newpage

\appendix

\section{Case Study for Synthesis QA}\label{sec:appendix-case}
\begin{promptbox}[Good Case 1]{lightgreen}
\textbf{Scientific Seed Snippet Texts:} \\
Given a system of oppositely charged layers, such as a double layer of ions in an electrolyte solution, explain why these layers do not combine (neutralize) with each other.\\
\textbf{Synthesis QA:} 

[Problem]
Given a system of oppositely charged layers, such as a double layer of ions in an electrolyte solution, explain why these layers do not combine (neutralize) with each other.
\\

[Solution]
In a system of oppositely charged layers, such as a double layer of ions in an electrolyte solution, the layers do not combine (neutralize) due to the electrostatic repulsion between the ions. This repulsion arises from the Coulombic force, which is a fundamental force in nature that acts between charged particles.
\\

To understand this, let's consider a simple example of a positively charged cation (e.g., Na+) and a negatively charged anion (e.g., Cl-). When these ions are brought close together, they experience an electrostatic force that pushes them apart. This force can be calculated using Coulomb's law, which states that the electrostatic force (F) between two point charges (q1 and q2) separated by a distance (r) is given by:$F = k * (q1 * q2) / r^2$, where k is the Coulomb constant (approximately $8.99 \times 10^9 N m^2 C^{-2}$).
\\

In the case of an electrolyte solution, the ions are surrounded by a cloud of counter-ions (ions of opposite charge) that neutralize their charge locally. This cloud of counter-ions creates an electric double layer around each ion, which prevents the oppositely charged ions from coming too close to each other and neutralizing. The repulsion between these double layers is known as the electrostatic double layer repulsion.
\\

The thickness of the double layer is typically on the order of a few angstroms (Å), and the strength of the repulsion decreases rapidly as the distance between the layers increases. This repulsion is responsible for the stability of colloidal suspensions, the behavior of charged surfaces in contact with electrolyte solutions, and many other phenomena in surface chemistry and electrochemistry.
\\

In summary, the electrostatic repulsion between oppositely charged layers in a system, such as the double layer of ions in an electrolyte solution, prevents the layers from combining (neutralizing) with each other due to the Coulombic force. This repulsion arises from the electric double layer around each ion, which is created by the counter-ions that neutralize the charge locally. The thickness of the double layer and the strength of the repulsion depend on factors such as the ionic strength of the solution, the surface charge density, and the dielectric constant of the medium.
\end{promptbox}

\section{Prompt}\label{sec:appendix-prompt}
\begin{promptbox}[Prompt for Scientific QA Synthesis]{lightblue}
\textbf{Instruction}

Please gain inspiration from the following \{Discipline Placeholder\} content to create a high-quality \{Discipline Placeholder\} problem and solution. Present your output in two distinct sections: [Problem] and [Solution].\\

\textbf{\{Discipline Placeholder\} Content}

\{Seed Snippet Placeholder\}\\

\textbf{Guidelines}

[Problem]: This should be **completely self-contained**, providing all the contextual information one needs to understand and solve the problem.\\

[Solution]: Present a comprehensive, step-by-step solution that solves the problem **correctly** and educates the student, around 250-350 words long. Clearly articulate the reasoning and methods used at each step, providing insight into the problem-solving process. Take care to format any equations properly using LaTeX or appropriate notation.\\
\end{promptbox}

\begin{promptbox}[Prompt for Topic Labeling]{lightblue}
I am categorizing a series of articles according to the following 11 topics. Next, I will give you an article, please select only one topic that the article is the most related to:\\

[Topics]: \{Topic List Placeholder\}\\

[Article]: \{Web Page Content Placeholder\}\\

Please only return the most related topic:
\end{promptbox}

\section{Example for Accuracy Degradation Transformations}\label{sec:appendix-corrupt-example}
 \emph{\textbf{\#\# Before Transformations:} In the given chemical reaction, we have sodium (Na) reacting with \textcolor{red}{chlorine} (Cl2) to form sodium chloride (NaCl). To determine the number of atoms of \textcolor{red}{chlorine} before and after the reaction, we will first count the number of \textcolor{red}{chlorine} atoms…adjust the coefficients of the reactants to make the number of \textcolor{red}{chlorine} atoms \textcolor{red}{equal} before and after the reaction:\textcolor{red}{2}Na + Cl\textcolor{red}{2} == \textcolor{red}{2}NaCl.}

 \emph{\textbf{\#\# After Transformations:} In the given chemical reaction, we have sodium (Na) reacting with \textcolor{red}{oxygen} (Cl2) to form sodium chloride (NaCl). To determine the number of atoms of \textcolor{red}{oxygen} before and after the reaction, we will first count the number of \textcolor{red}{oxygen} atoms…adjust the coefficients of the reactants to make the number of \textcolor{red}{oxygen} atoms \textcolor{red}{unequal} before and after the reaction:\textcolor{red}{6}Na + Cl\textcolor{red}{3} == \textcolor{red}{8}NaCl}

 In this example, "chlorine" is replaced with a random hyponym (oxygen, hydrogen, neon, etc.) of its hypernym (chemical element), the numbers in the chemical formulas are randomly replaced, and the adjective "equal" is replaced with "unequal."

\section{Detailed Surrogate Experiment Results}\label{sec:appendix-tiny-detail}

When introducing surrogate experiments with TinyLlama in Section~\ref{surrogate_experiment}, we select several representative benchmarks for computing the average performance to avoid large performance discrepancies across benchmarks. Here we report all benchmark results. ``PHY'', ``CHE'', and ``BIO'' denote the physics, chemistry, and biology sub-tasks of the corresponding benchmarks. The best and second best are in \textbf{bold} and \underline{underlined}, respectively.

 \begin{table*}[t]
    \centering
    \small
    \caption{Few-shot performance of TinyLlama continually pre-trained on different corpora on major benchmarks.}
    \setlength\tabcolsep{0.75mm}{
    \begin{tabular}{lccccccccccc}
    \toprule
    \multirow{2.5}{*}{\textbf{Models}} & \multicolumn{3}{c}{\textbf{Bilingual}} & \multicolumn{5}{c}{\textbf{Math}} & \multicolumn{2}{c}{\textbf{Code}} \\
    \cmidrule(lr){2-4}\cmidrule(lr){5-9}\cmidrule(lr){10-11}
     & \textbf{MMLU} & \textbf{C-Eval} & \textbf{CMMLU} & \textbf{MATH} & \textbf{GSM8K} & \textbf{ASDiv} & \textbf{MAWPS} & \textbf{SAT-Math} & \textbf{HumanEval} & \textbf{MBPP} &  \\
    \midrule[0.5pt]
    TinyLlama & 25.70 & 25.11 & 25.09 & 2.80 & \underline{3.00} & 18.00 & 20.30 & 23.64 & \textbf{10.37} & \textbf{13.40} \\
    \quad \emph{w/} 5B (1B Norm.) & \underline{28.35} & \underline{30.02} & \underline{29.10} & \underline{2.90} & 2.00 & \underline{21.00} & \underline{31.40} & \textbf{24.09} & 4.88 & 4.60 \\
    \quad \emph{w/} 5B (1B Syn.) & \textbf{31.89} & \textbf{34.60} & \textbf{35.09} & \textbf{5.30} & \textbf{14.90} & \textbf{48.10} & \textbf{66.40} & \underline{23.65} & \underline{9.15} & \underline{6.80} \\
    \bottomrule
    \end{tabular}}
    \label{tab:ablation_web_general}
\end{table*}

\begin{table*}[t]
    \centering
    \small
    \caption{Few-shot performance of TinyLlama continually pre-trained on different corpora on scientific benchmarks.}
    \setlength\tabcolsep{1.0mm}{
    \begin{tabular}{lcccccccccccc}
    \toprule
    \multirow{2.5}*{\textbf{Models}} & \multicolumn{4}{c}{\textbf{SciEval}} & \multicolumn{1}{c}{\textbf{SciQ}} & \multicolumn{3}{c}{\textbf{GaoKao}} & \multicolumn{3}{c}{\textbf{ARC}} & \multicolumn{1}{c}{\textbf{AQUA-RAT}} \\
    \cmidrule(lr){2-5}\cmidrule(lr){6-6}\cmidrule(lr){7-9}\cmidrule(lr){10-12}\cmidrule(lr){13-13}
     & {PHY} & {CHE} & {BIO} & {Avg.} & {Avg.} & {MathQA} & {CHE} & {BIO} & {Easy} & {Challenge} & {Avg.} & {Avg.} \\
    \midrule[0.5pt]
    TinyLlama &  26.22 & 27.22 & 31.94 & 28.85 & 24.60 & 22.79 & \textbf{27.05} & 20.00 & 24.87 & 26.19 & 25.53 & \underline{22.05}\\
    \quad \emph{w/} 5B (1B Norm.) & \underline{28.32} & \underline{35.64} & \underline{45.62} & \underline{38.64} & \underline{56.10} & \underline{26.50} & \textbf{27.05} & \textbf{30.48} & \underline{37.75} & \underline{30.55} & \underline{34.15} & \textbf{24.02} \\
    \quad \emph{w/} 5B (1B Syn.) & \textbf{31.10} & \textbf{38.26} & \textbf{47.81} & \textbf{40.90} & \textbf{60.30} & \textbf{27.35} & \textbf{27.05} & \underline{29.52} & \textbf{45.45} & \textbf{34.13} & \textbf{39.79} & 20.87 \\
    \bottomrule
    \end{tabular}}
    \label{tab:ablation_web_science}
\end{table*}

\begin{table*}[t]
    \centering
    \small
    \caption{Few-shot performance of TinyLlama continually pre-trained on varying corruption levels of synthetic data on major benchmarks.}
    \setlength\tabcolsep{1.3mm}{
    \begin{tabular}{lccccccccccc}
    \toprule
    \multirow{2.5}{*}{\textbf{Models}} & \multicolumn{3}{c}{\textbf{Bilingual}} & \multicolumn{5}{c}{\textbf{Math}} & \multicolumn{2}{c}{\textbf{Code}} \\
    \cmidrule(lr){2-4}\cmidrule(lr){5-9}\cmidrule(lr){10-11}
     & \textbf{MMLU} & \textbf{C-Eval} & \textbf{CMMLU} & \textbf{MATH} & \textbf{GSM8K} & \textbf{ASDiv} & \textbf{MAWPS} & \textbf{SAT-Math} & \textbf{HumanEval} & \textbf{MBPP} &  \\
    \midrule[0.5pt]
    TinyLlama & 25.70 & 25.11 & 25.09 & 2.80 & 3.00 & 18.00 & 20.30 & 23.64 & \underline{10.37} & \textbf{13.40} \\
    \quad \emph{w/} 0.0 & \underline{31.89} & \textbf{34.60} & \textbf{35.09} & \textbf{5.30} & 14.90 & \underline{48.10} & \underline{66.40} & 23.64 & 9.15 & 6.80 \\
    \quad \emph{w/} 0.3 & 31.28 & \underline{31.94} & \underline{34.08} & \textbf{5.30} & 15.50 & \textbf{49.00} & 65.60 & \textbf{24.55} & \textbf{10.98} & 7.60 \\
    \quad \emph{w/} 0.4 & \textbf{32.54} & 31.67 & 33.79 & 4.60 & 10.50 & 37.50 & 57.50 & 23.64 & 9.15 & 8.60 \\
    \quad \emph{w/} 0.5 & 30.23 & 31.27 & 33.44 & \underline{4.90} & \underline{15.80} & 47.60 & 64.90 & 22.73 & \textbf{10.98} & 8.60 \\
    \quad \emph{w/} 0.6 & 28.22 & 29.87 & 33.00 & 4.60 & \textbf{16.90} & 47.90 & \textbf{67.40} & 23.18 & 8.54 & \underline{9.60} \\
    \quad \emph{w/} 0.7 & 27.65 & 27.73 & 32.30 & 4.80 & 1.00 & 4.50 & 3.70 & \underline{24.09} & 9.76 & 8.80 \\
    \bottomrule
    \end{tabular}}
    \label{tab:synth_acc_general}
\end{table*}

\begin{table*}[t]
    \centering
    \small
    \caption{Few-shot performance of TinyLlama continually pre-trained on varying corruption levels of synthetic data on scientific benchmarks.}
    \setlength\tabcolsep{1.5mm}{
    \begin{tabular}{lcccccccccccc}
    \toprule
    \multirow{2.5}*{\textbf{Models}} & \multicolumn{4}{c}{\textbf{SciEval}} & \multicolumn{1}{c}{\textbf{SciQ}} & \multicolumn{3}{c}{\textbf{GaoKao}} & \multicolumn{3}{c}{\textbf{ARC}} & \multicolumn{1}{c}{\textbf{AQUA-RAT}} \\
    \cmidrule(lr){2-5}\cmidrule(lr){6-6}\cmidrule(lr){7-9}\cmidrule(lr){10-12}\cmidrule(lr){13-13}
     & {PHY} & {CHE} & {BIO} & {Avg.} & {Avg.} & {MathQA} & {CHE} & {BIO} & {Easy} & {Challenge} & {Avg.} & {Avg.} \\
    \midrule[0.5pt]
    TinyLlama & 26.22 & 27.22 & 31.94 & 28.85 & 24.60 & 22.79 & \underline{27.05} & 20.00 & 24.87 & 26.19 & 25.53 & \textbf{22.05} \\
    \quad \emph{w/} 0.0 & 31.10 & \underline{38.26} & \underline{47.81} & 40.90 & \underline{60.30} & \textbf{27.35} & \underline{27.05} & \underline{29.52} & \textbf{45.45} & 34.13 & 39.79 & 20.87 \\
    \quad \emph{w/} 0.3 & \underline{36.59} & 37.64 & \textbf{48.23} & \underline{41.45} & \textbf{60.80} & 22.79 & \underline{27.05} & 21.43 & 43.06 & 32.94 & 38.00 & \underline{21.26} \\
    \quad \emph{w/} 0.4 & \textbf{38.41} & \textbf{39.19} & 46.76 & \textbf{41.91} & 57.20 & \underline{23.36} & 22.22 & 27.14 & \underline{45.37} & \textbf{36.43} & \textbf{40.90} & 19.69  \\
    \quad \emph{w/} 0.5 & 34.15 & 37.79 & 43.01 & 39.27 & 58.10 & \underline{23.36} & \textbf{27.54} & \textbf{32.86} & 44.95 & \underline{35.41} & \underline{40.18} & 20.47  \\
    \quad \emph{w/} 0.6 & 34.15 & 35.46 & 44.26 & 38.57 & 50.10 & 22.51 & 26.09 & 26.67 & 40.91 & 31.23 & 36.07 & 17.32  \\
    \quad \emph{w/} 0.7 & 33.54 & 31.88 & 43.63 & 36.47 & 50.50 & 22.51 & 26.57 & 24.29 & 40.57 & 30.38 & 35.47 & 18.11 \\
    \bottomrule
    \end{tabular}}
    \label{tab:synth_acc_science}
\end{table*}

\begin{table*}[t]
    \centering
    \small
    \caption{Few-shot performance of TinyLlama after training with different ratios of synthetic data on major benchmarks.}
    \setlength\tabcolsep{1.3mm}{
    \begin{tabular}{lccccccccccc}
    \toprule
    \multirow{2.5}{*}{\textbf{Models}} & \multicolumn{3}{c}{\textbf{Bilingual}} & \multicolumn{5}{c}{\textbf{Math}} & \multicolumn{2}{c}{\textbf{Code}} \\
    \cmidrule(lr){2-4}\cmidrule(lr){5-9}\cmidrule(lr){10-11}
     & \textbf{MMLU} & \textbf{C-Eval} & \textbf{CMMLU} & \textbf{MATH} & \textbf{GSM8K} & \textbf{ASDiv} & \textbf{MAWPS} & \textbf{SAT-Math} & \textbf{HumanEval} & \textbf{MBPP} &  \\
    \midrule[0.5pt]
    TinyLlama & 25.70 & 25.11 & 25.09 & 2.80 & 3.00 & 18.00 & 20.30 & \underline{23.64} & \underline{10.37} & \textbf{13.40} \\
    \quad \emph{w/} 1:10 & 25.73 & 28.58 & 32.94 & 4.90 & 5.20 & 9.40 & 16.10 & \textbf{27.27} & 8.54 & 8.20 \\
    \quad \emph{w/} 2:10 & \textbf{31.89} & \textbf{34.60} & \textbf{35.09} & 5.30 & \underline{14.90} & \underline{48.10} & \textbf{66.40} & \underline{23.64} & 9.15 & 6.80 \\
    \quad \emph{w/} 3:10 & 27.62 & \underline{32.25} & 33.31 & \textbf{6.60} & 2.20 & 20.90 & 30.10 & 22.73 & \textbf{10.98} & \underline{8.60} \\
    \quad \emph{w/} 4:10 & \underline{30.25} & 29.43 & \underline{34.36} & \underline{5.60} & \textbf{15.50} & \textbf{50.40} & \underline{64.90} & 22.60 & 7.32 & 8.40 \\
    \bottomrule
    \end{tabular}}
    \label{tab:science-ratio-general}
\end{table*}

\begin{table*}[t]
    \centering
    \small
    \caption{Few-shot performance of TinyLlama after training with different ratios of synthetic data on scientific benchmarks.}
    \setlength\tabcolsep{1.5mm}{
    \begin{tabular}{lcccccccccccc}
    \toprule
    \multirow{2.5}*{\textbf{Models}} & \multicolumn{4}{c}{\textbf{SciEval}} & \multicolumn{1}{c}{\textbf{SciQ}} & \multicolumn{3}{c}{\textbf{GaoKao}} & \multicolumn{3}{c}{\textbf{ARC}} & \multicolumn{1}{c}{\textbf{AQUA-RAT}} \\
    \cmidrule(lr){2-5}\cmidrule(lr){6-6}\cmidrule(lr){7-9}\cmidrule(lr){10-12}\cmidrule(lr){13-13}
     & {PHY} & {CHE} & {BIO} & {Avg.} & {Avg.} & {MathQA} & {CHE} & {BIO} & {Easy} & {Challenge} & {Avg.} & {Avg.} \\
    \midrule[0.5pt]
    TinyLlama & 26.22 & 27.22 & 31.94 & 28.85 & 24.60 & \underline{22.79} & \textbf{27.05} & 20.00 & 24.87 & 26.19 & 25.53 & \textbf{22.05} \\
    \quad \emph{w/} 1:10 & \textbf{36.59} & 34.53 & 42.17 & 37.64 & 50.10 & \underline{22.79} & \textbf{27.05} & \underline{24.76} & 39.69 & 32.59 & 36.14 & 19.69 \\
    \quad \emph{w/} 2:10 & \underline{31.10} & \textbf{38.26} & \textbf{47.81} & \textbf{40.90} & \textbf{60.30} & \textbf{27.35} & \textbf{27.05} & \textbf{29.52} & \underline{45.45} & \underline{34.13} & \textbf{39.79} & 20.87 \\
    \quad \emph{w/} 3:10 & 27.80 & \underline{37.79} & \underline{46.35} & 37.98 & \underline{58.00} & \underline{22.79} & \underline{26.57} & 21.43 & 44.57 & 33.70 & 39.14 & \underline{21.65} \\
    \quad \emph{w/} 4:10 & 29.88 & 36.39 & 43.84 & \underline{38.34} & 57.20 & \underline{22.79} & \textbf{27.05} & 20.00 & \textbf{48.57} & \textbf{36.86} & \underline{39.71} & 19.04 \\
\bottomrule
    \end{tabular}
    }
    \label{tab:science-ratio}
\end{table*}

\begin{table*}[t]
    \centering
    \small
    \caption{Few-shot performance of TinyLlama with different data curriculum methods on major benchmarks.}
    \setlength\tabcolsep{1.2mm}{
    \begin{tabular}{lccccccccccc}
    \toprule
    \multirow{2.5}{*}{\textbf{Models}} & \multicolumn{3}{c}{\textbf{Bilingual}} & \multicolumn{5}{c}{\textbf{Math}} & \multicolumn{2}{c}{\textbf{Code}} \\
    \cmidrule(lr){2-4}\cmidrule(lr){5-9}\cmidrule(lr){10-11}
     & \textbf{MMLU} & \textbf{C-Eval} & \textbf{CMMLU} & \textbf{MATH} & \textbf{GSM8K} & \textbf{ASDiv} & \textbf{MAWPS} & \textbf{SAT-Math} & \textbf{HumanEval} & \textbf{MBPP} \\
    \midrule[0.5pt]
    TinyLlama & 25.70 & 25.11 & 25.09 & 2.80 & 3.00 & 18.00 & 20.30 & 23.64 & \underline{10.37} & \textbf{13.40} \\
    \quad \emph{w/} RM & \underline{31.89} & \underline{34.60} & \underline{35.09} & \underline{5.30} & \underline{14.90} & \underline{48.10} & \underline{66.40} & 23.65 & 9.15 & 6.80 \\
    \quad \emph{w/} PB & 26.78 & 23.73 & 27.58 & 3.50 & 6.10 & 36.60 & 45.50 & 24.09 & 6.71 & 7.80 \\
    \quad \emph{w/} BP & 26.98 & 24.14 & 28.63 & 3.80 & 5.00 & 32.20 & 43.40 & 23.18 & 6.71 & 8.00 \\
    \quad \emph{w/} PBP & 26.86 & 24.15 & 27.59 & 2.90 & 7.00 & 36.30 & 46.20 & 24.55 & 6.10 & 6.20 \\
    \quad \emph{w/} HL & 27.78 & 30.49 & 32.24 & 4.10 & 10.50 & 38.80 & 58.30 & \underline{25.91} & 8.54 & \underline{11.20} \\
    \quad \emph{w/} LH & \textbf{32.16} & \textbf{36.89} & \textbf{37.27} & \textbf{6.10} & \textbf{20.60} & \textbf{53.90} & \textbf{70.80} & \textbf{26.36} & \textbf{12.80} & 8.80 \\
    \bottomrule
    \end{tabular}}
    \label{tab:curriculum-general}
\end{table*}

\begin{table*}[t]
    \centering
    \small
    \caption{Few-shot performance of TinyLlama with different data curriculum methods on scientific benchmarks.}
    \setlength\tabcolsep{1.4mm}{
    \begin{tabular}{lcccccccccccc}
    \toprule
    \multirow{2.5}*{\textbf{Models}} & \multicolumn{4}{c}{\textbf{SciEval}} & \multicolumn{1}{c}{\textbf{SciQ}} & \multicolumn{3}{c}{\textbf{GaoKao}} & \multicolumn{3}{c}{\textbf{ARC}} & \multicolumn{1}{c}{\textbf{AQUA-RAT}} \\
    \cmidrule(lr){2-5}\cmidrule(lr){6-6}\cmidrule(lr){7-9}\cmidrule(lr){10-12}\cmidrule(lr){13-13}
     & {PHY} & {CHE} & {BIO} & {Avg.} & {Avg.} & {MathQA} & {CHE} & {BIO} & {Easy} & {Challenge} & {Avg.} & {Avg.} \\
    \midrule[0.5pt]
    TinyLlama & 26.22 & 27.22 & 31.94 & 28.85 & 24.60 & 22.79 & \textbf{27.05} & 20.00 & 24.87 & 26.19 & 25.53 & 22.05 \\
    \quad \emph{w/} RM & 31.10 & \underline{38.26} & \underline{47.81} & \underline{40.90} & \underline{60.30} & \underline{27.35} & \textbf{27.05} & 29.52 & 45.45 & 34.13 & 39.79 & 20.87 \\
    \quad \emph{w/} PB & 32.32 & 32.04 & 41.54 & 35.61 & 35.10 & \textbf{29.34} & \underline{26.57} & \underline{31.90} & 36.74 & 28.75 & 32.75 & \textbf{25.20} \\
    \quad \emph{w/} BP &	31.10 & 33.90 & 42.59 & 36.78 & 46.90 & 22.51 & 23.19 & 29.52 & 36.15 & 30.29 & 33.22 & \underline{24.02} \\
    \quad \emph{w/} PBP & 30.49 & 34.53 & 41.96 & 36.78 & 49.60 & \underline{27.35} & 24.64 & \textbf{32.86} & 45.88 & 32.68 & 39.28 & 20.08 \\
    \quad \emph{w/} HL & \underline{32.93} & 34.06 & 43.84 & 37.56 & 55.20 & 22.51 & \underline{26.57} & 31.43 & \underline{50.72} & \underline{38.14} & \underline{44.43} & 23.23 \\
    \quad \emph{w/} LH & \textbf{37.20} & \textbf{41.84} & \textbf{51.15} & \textbf{44.71} & \textbf{65.50} & 25.07 & 26.09 & 22.38 & \textbf{57.62} & \textbf{41.81} & \textbf{49.71} & 18.50 \\
\bottomrule
    \end{tabular}
    }
    \label{tab:curriculum-science}
\end{table*}

\begin{table*}[t]
    \centering
    \small
    \caption{Few-shot performance of TinyLlama continually pre-trained on different open-source datasets on major benchmarks.}
    \setlength\tabcolsep{0.7mm}{
    \begin{tabular}{lccccccccccc}
    \toprule
    \multirow{2.5}{*}{\textbf{Models}} & \multicolumn{3}{c}{\textbf{Bilingual}} & \multicolumn{5}{c}{\textbf{Math}} & \multicolumn{2}{c}{\textbf{Code}} \\
    \cmidrule(lr){2-4}\cmidrule(lr){5-9}\cmidrule(lr){10-11}
     & \textbf{MMLU} & \textbf{C-Eval} & \textbf{CMMLU} & \textbf{MATH} & \textbf{GSM8K} & \textbf{ASDiv} & \textbf{MAWPS} & \textbf{SAT-Math} & \textbf{HumanEval} & \textbf{MBPP} &  \\
    \midrule[0.5pt]
    TinyLlama & 25.70 & 25.11 & 25.09 & 2.80 & 3.00 & 18.00 & 20.30 & \underline{23.64} & \textbf{10.37} & \textbf{13.40} \\
    \quad \emph{w/} 5B (1B WebIns.) & 26.85 & \underline{32.73} & \underline{33.22} & \textbf{7.50} & 0.80 & 1.80 & 2.40 & \textbf{25.00} & 6.71 & 5.20 \\
    \quad \emph{w/} 5B (1B Cosm.) & \underline{27.51} & 28.08 & 31.51 & \underline{6.90} & \textbf{19.90} & \textbf{49.70} & \textbf{68.20} & 23.18 & \underline{9.15} & \underline{7.40} \\
    \quad \emph{w/} 5B (1B Syn.) & \textbf{31.89} & \textbf{34.60} & \textbf{35.09} & 5.30 & \underline{14.90} & \underline{48.10} & \underline{66.40} & \underline{23.64} & \underline{9.15} & 6.80 \\
    \bottomrule
    \end{tabular}}
    \label{tab:compare-dataset-general}
\end{table*}

\begin{table*}[t]
    \centering
    \small
    \caption{Few-shot performance of TinyLlama continually pre-trained on different open-source datasets on scientific benchmarks.}
    \setlength\tabcolsep{1mm}{
    \begin{tabular}{lcccccccccccc}
    \toprule
    \multirow{2.5}*{\textbf{Models}} & \multicolumn{4}{c}{\textbf{SciEval}} & \multicolumn{1}{c}{\textbf{SciQ}} & \multicolumn{3}{c}{\textbf{GaoKao}} & \multicolumn{3}{c}{\textbf{ARC}} & \multicolumn{1}{c}{\textbf{AQUA-RAT}} \\
    \cmidrule(lr){2-5}\cmidrule(lr){6-6}\cmidrule(lr){7-9}\cmidrule(lr){10-12}\cmidrule(lr){13-13}
     & {PHY} & {CHE} & {BIO} & {Avg.} & {Avg.} & {MathQA} & {CHE} & {BIO} & {Easy} & {Challenge} & {Avg.} & {Avg.} \\
    \midrule[0.5pt]
    TinyLlama & 26.22 & 27.22 & 31.94 & 28.85 & 24.60 & 22.79 & \textbf{27.05} & 20.00 & 24.87 & 26.19 & 25.53 & \underline{22.05}\\
    \quad \emph{w/} 5B (1B WebIns.) & \underline{32.32} & 34.21 & \underline{44.26} & 37.71 & \underline{47.70} & 23.36 & \textbf{27.05} & \textbf{31.90} & 36.36 & 32.94 & 34.65 & 20.87 \\
    \quad \emph{w/} 5B (1B Cosm.) & \textbf{34.76} & \underline{35.77} & \underline{44.26} & \underline{38.80} & 41.30 & \underline{26.21} & \underline{25.60} & 27.62 & \underline{43.81} & \textbf{36.95} & \textbf{40.38} & \textbf{22.83}\\
    \quad \emph{w/} 5B (1B Syn.) & 31.10 & \textbf{38.26} & \textbf{47.81} & \textbf{40.90} & \textbf{60.30} & \textbf{27.35} & \textbf{27.05} & \underline{29.52} & \textbf{45.45} & \underline{34.13} & \underline{39.79} & 20.87 \\ 
\bottomrule
    \end{tabular}
    }
    \label{tab:compare-dataset-science}
\end{table*}

\end{document}